%% file: main.tex
\documentclass{article}

\usepackage[bottom]{footmisc}
\usepackage{microtype}
\usepackage{graphicx}
\usepackage{booktabs} % for professional tables
\usepackage{inconsolata}

\usepackage{hyperref}

\usepackage{times}
\usepackage{latexsym}
\usepackage{float}

\usepackage[T1]{fontenc}
\usepackage[utf8]{inputenc}

\usepackage[accepted]{icml2023}

\usepackage{amsmath}
\usepackage{amssymb}
\usepackage{mathtools}
\usepackage{amsthm}
\usepackage{url}

\usepackage[capitalize,noabbrev]{cleveref}

\theoremstyle{plain}

\theoremstyle{definition}

\theoremstyle{remark}

\usepackage[textsize=tiny]{todonotes}

\input{packages_and_commands}

\icmltitlerunning{\semsupxc{}: Semantic Supervision for Zero and Few-shot Extreme Classification}

\begin{document}
% \raggedbottom

\twocolumn[
% \icmltitle{\semsupxc{}: Semantic Supervision for Extreme Classification}
% \icmltitle{\semsupxc{}: Improving Generalization for Extreme Classification}
\icmltitle{\semsupxc{}: Semantic Supervision for Zero and Few-shot \\  Extreme Classification}



\icmlsetsymbol{equal}{*}

\begin{icmlauthorlist}
\icmlauthor{Pranjal Aggarwal}{prn}
\icmlauthor{Ameet Deshpande}{prince}
\icmlauthor{Karthik Narasimhan}{prince}
\end{icmlauthorlist}
% \begin{center}
% \textsuperscript{1}Indian Institute of Technology, Delhi\\ \textsuperscript{2}Department of Computer Science, Princeton University \\
%  \texttt{pranjal2041@gmail.com}, \texttt{\{asd, karthikn\}@cs.princeton.edu}
%  \end{center}
 \vspace{-2pt}

\icmlaffiliation{prn}{Department of Computer Science and Engineering, Indian Institute of Technology, Delhi, India}
\icmlaffiliation{prince}{Department of Computer Science, Princeton University}
% \icmlaffiliation{krn}{Department of Computer Science, Princeton University}

\icmlcorrespondingauthor{Pranjal Aggarwal}{pranjal2041@gmail.com}
\icmlcorrespondingauthor{Ameet Deshapande, Karthik Narasimhan}{\{ asd, karthik\}@cs.princeton.edu}
% \icmlcorrespondingauthor{Karthik Narasimhan}{karthikn@cs.princeton.edu}

% \author{
% \textbf{Pranjal Aggarwal\textsuperscript{1}}, \textbf{Ameet Deshpande\textsuperscript{2}}, \textbf{Karthik Narasimhan\textsuperscript{2}}\\
% \textsuperscript{1}Indian Institute of Technology, Delhi\\ \textsuperscript{2}Department of Computer Science, Princeton University \\
%  \texttt{pranjal2041@gmail.com}, \texttt{\{asd, karthikn\}@cs.princeton.edu}
% }

\icmlkeywords{Machine Learning, NLP, Extreme Classification}

\vskip 0.3in
]



% \Notice@String
\printAffiliationsAndNotice{}  % leave blank if no need to mention equal contribution
% \title{\semsupxc{}: Semantic Supervision for Extreme Classification}

% \maketitle

%%%%%%%%%%%%%%%%%%%%%%%%%%%%%%%%%%%%%%%%%%%%%%%%%%%%%%%%%%%%%%%%%%%%%%%%%%%%%%%
\input{LaTeX/000_files}

\section*{Acknowledgements}
This material is based upon work supported by the National Science Foundation under Grant No. 2239363. Any opinions, findings, and conclusions or recommendations expressed in this material are those of the author(s) and do not necessarily reflect the views of the National Science Foundation. We also acknowledge support from the Chadha Center for Global India at Princeton University.
We thank Jens Tuyls, Khanh Nguyen and other members of the Princeton-NLP group for useful feedback and comments.

% \clearpage

\bibliography{iclr2023_conference}

\bibliographystyle{icml2023}

\clearpage

\appendix

\input{LaTeX/900_appendix_a}

\end{document}

%% file: packages_and_commands.tex
\usepackage{amsmath,amsfonts}
\usepackage{booktabs,arydshln}
\usepackage{multirow}
\usepackage{ulem}
\usepackage{pifont}
\usepackage{relsize}
\usepackage{enumitem}
\usepackage{graphics}
\usepackage{mathtools}
\usepackage{amsmath,amsfonts,amssymb}
\usepackage{colortbl} 
\usepackage{subcaption}
\usepackage{tabularx}
\usepackage{arydshln}
\usepackage{breqn}
\usepackage{graphicx}
\usepackage{xcolor, soul}
\definecolor{highlightcolor}{HTML}{d4fcfb}
\sethlcolor{highlightcolor}
\makeatletter
\def\adl@drawiv#1#2#3{%
        \hskip.5\tabcolsep
        \xleaders#3{#2.5\@tempdimb #1{1}#2.5\@tempdimb}%
                #2\z@ plus1fil minus1fil\relax
        \hskip.5\tabcolsep}
\newcommand{\cdashlinelr}[1]{%
  \noalign{\vskip\aboverulesep
           \global\let\@dashdrawstore\adl@draw
           \global\let\adl@draw\adl@drawiv}
  \cdashline{#1}
  \noalign{\global\let\adl@draw\@dashdrawstore
           \vskip\belowrulesep}}
\makeatother

\newcommand{\cmmnt}[1]{\ignorespaces}

\newcommand{\symbolsecref}[1]{($\S$~\ref{#1})}

\newcommand{\semsup}[0]{\textsc{semsup}}
\newcommand{\semsupxc}[0]{SemSup-XC}
\newcommand{\semsupxcbig}[0]{SEMSUP-XC}
\newcommand{\semsupweb}[0]{\textsc{semsup-web}}
\newcommand{\xc}[0]{XC}
\newcommand{\maclr}[0]{MACLR}
\newcommand{\maclrdescs}[0]{MACLR with Descriptions}
\newcommand{\zest}[0]{ZestXML}
\newcommand{\groov}[0]{GROOV}
\newcommand{\groovdescs}[0]{GROOV with Descriptions}

\newcommand{\lightxml}[0]{Light XML}
\newcommand{\tfidf}[0]{TF-IDF}
\newcommand{\tfive}[0]{T5}
\newcommand{\senttrans}[0]{Sentence-Transformer}

\newcommand{\coil}[0]{COIL}
\newcommand{\relaxed}[0]{Hybrid-Match}

\newcommand{\zs}[0]{ZS}
\newcommand{\gzs}[0]{G-ZS}
\newcommand{\fs}[0]{FS}
\newcommand{\zsxc}[0]{\zs{}-\xc{}}
\newcommand{\gzsxc}[0]{GZS-\xc{}}
\newcommand{\fsxc}[0]{\fs{}-\xc{}}

\newcommand{\precision}[1]{P@#1}
\newcommand{\recall}[1]{R@#1}

\newcommand{\finalnum}[1]{{#1}}
\newcommand{\tablestd}[1]{{\tiny \textcolor{gray}{$(\pm #1)$}}}

\newcommand{\inputencoder}[0]{$f_{\tiny \textrm{IE}}$}
\newcommand{\outputencoder}[0]{$g_{\tiny \textrm{OE}}$}

\newcommand{\clstoken}[0]{\texttt{[CLS]}}
\newcommand{\amazon}[0]{AmazonCat}
\newcommand{\eurlex}[0]{EURLex}
\newcommand{\wikipedia}[0]{Wikipedia}

\newcommand{\amazonfull}[0]{AmazonCat-13K}
\newcommand{\eurlexfull}[0]{EURLex-4.3K}
\newcommand{\wikipediafull}[0]{Wikipedia-1M}

\renewcommand{\cite}[1]{\citep{#1}}

\definecolor{mygreen}{RGB}{0,160,0}
\newcommand{\cmarkgreen}{\textcolor{mygreen}{{\Large \ding{51}}}}
\definecolor{myred}{RGB}{178,34,34}
\newcommand{\xmarkred}{\textcolor{myred}{{\Large \ding{55}}}}

\definecolor{BlueGreen}{rgb}{0.0, 0.87, 0.87}
\definecolor{WildStrawberry}{rgb}{1.0, 0.26, 0.64}
\definecolor{approach}{HTML}{a5daf3}
\newcommand*\circled[4]{\tikz[baseline=(char.base)]{
    \node[shape=circle, fill=#2, draw=#3, text=#4, inner sep=1.2pt, scale=0.9, font=\bfseries] (char) {#1};}}

\definecolor{darkyellowtable}{HTML}{806b03}
\definecolor{darkredtable}{HTML}{730202}

%% file: LaTeX/000_files.tex
% Abstract
\input{LaTeX/100_abstract}

% Introduction
\input{LaTeX/200_introduction}

\input{LaTeX/400_methods}
\input{LaTeX/500_experimental_setup}

\input{LaTeX/600_results}

\input{LaTeX/700_ablations}
\input{LaTeX/300_related_work}
\input{LaTeX/750_computation}

\input{LaTeX/800_conclusion}

\input{LaTeX/850_limitations}

%% file: LaTeX/100_abstract.tex
\begin{abstract}
% \vspace{-2pt}
Extreme classification (\xc{}) involves predicting over large numbers of classes (thousands to millions), with real-world applications like news article classification and e-commerce product tagging.
% , serving search engine results, and news article classification.
The zero-shot version of this task requires generalization to novel classes without additional supervision.
% , like a new class \textit{``fidget spinner''} for e-commerce product tagging.
% Applications involving large label spaces are often accompanied with addition of new categories, e.g., \textit{``fidget spinner''} in e-commerce product tagging, making generalization to novel classes crucial (zero-shot and few-shot classification).
In this paper, we develop \semsupxc{}, a model that achieves state-of-the-art zero-shot and few-shot performance on three \xc{} datasets derived from legal, e-commerce, and Wikipedia data.
% builds upon the recently proposed framework of semantic supervision that
To develop \semsupxc{}, we use automatically collected semantic class descriptions to represent classes and facilitate generalization through a novel hybrid matching module that matches input instances to class descriptions using a combination of semantic and lexical similarity.
% over contextual representations. % of similar tokens.
% incorporates  uses contrastive learning to train our proposed hybrid semantic-plus-lexical similarity module to match input instances to class descriptions, thus enabling ZS and FS classification.
% We train efficiently using contrastive learning over label descriptions associated with positive and negative classes.
% Specifically, we use a combination of contrastive learning, a hybrid lexico-semantic similarity module and automated description collection to train \semsupxc{} efficiently over extremely large class spaces.
% Our method, dubbed \semsupxc{}, is based on a recently proposed framework called semantic supervision (\semsup{}) which uses semantic label descriptions to represent and generalize to classes (e.g., ``A popular spinning toy intended as a stress reliever'' for \textit{``fidget spinner''}), but their method is computationally intractable for large label spaces.
% \semsupxc{} improves \semsup{}'s training time complexity (through contrastive learning), model performance (through hybrid lexical and semantic similarity between the instance and label description), and label description quality (through automated web scraping).
Trained with contrastive learning, \semsupxc{} significantly outperforms baselines and establishes state-of-the-art performance on all three datasets considered, gaining up to 12 precision points on zero-shot and more than 10 precision points on one-shot tests, with similar gains for recall@10.
Our ablation studies highlight the relative importance of our hybrid matching module %(up to 2 P@1 improvement on \amazon{}) 
and automatically collected class descriptions.% (up to 5 P@1 improvement on \amazon{}).
%  various components and conclude the combined importance of the proposed architecture and automatically scraped descriptions with improvements up to 33 precision@1 points. 
% Furthermore, qualitative analyses demonstrate \semsupxc{}'s better understanding of label space than other state-of-the-art models.  
% Through extensive experiments on three diverse datasets with large label spaces (4K, 13K, and 1M), we show that \semsupxc{} significantly outperforms baselines by 6 to 10 precision points (@1) on ZS classification and over 10 precision points on FS classification, with similar gains for recall@10 (3 for ZS and 2 for FS).
% \rebuttal{Our ablation studies show the relative importance of various components and conclude the combined importance of the proposed architecture and automatically scraped descriptions with improvements up to 33 precision@1 points. 
% Furthermore, qualitative analyses demonstrate \semsupxc{}'s better understanding of label space than other state-of-the-art models.  
% }
% Our ablation studies and qualitative analyses demonstrate the relative importance of our various improvements and show that~\semsupxc{}'s automated pipeline offers a consistently efficient method for extreme classification.
% \footnote{Code :  and demo are available at \url{https://github.com/princeton-nlp/semsup-xc} and \url{https://huggingface.co/spaces/Pranjal2041/SemSup-XC/}, respectively.}
\footnote{Code and demo are available at \url{https://github.com/princeton-nlp/semsup-xc} and
\url{https://huggingface.co/spaces/Pranjal2041/SemSup-XC/}.}

% \karthik{this last sentence could use some concrete finding.}
% We further show that access to simple data augmentation techniques can boost performance by over 2 points across the board.
\end{abstract}

%% file: LaTeX/200_introduction.tex
\section{Introduction}
\label{sec:intro}

% Teaser figure
\input{LaTeX/Figure_Latex/teaser_figure}

% \karthik{Outline:
% 1. \sout{Explain extreme classification, what is the challenge (large spaces) and why current approaches are not sufficient}
% 2. \sout{Need for semantic framing of the label space - this helps generalize better since models learn to look at features and aspects of a class rather than map directly ot a number.}
% 3. \sout{SemSUP - recently introduced framework that provides the above semantic space. However, there are issues that prevent naive use ...}
% 4. \sout{We solve these by .... (add technical contributions to improving semsup for XC).}
% 5. Key experimental findings, ablations, etc.
% 6. End with future challenges (since the numbers are still low) and discussion.
% }

% Siamese XML is a good paper to motivate the SemSup aspect
Extreme classification (\xc{}) studies the problem of predicting over a large space of classes, ranging from thousands to millions~\citep{Agrawal2013MultilabelLW, bengio2019extreme,Bhatia2015SparseLE, Chang2019AMD, Lin2014MultilabelCV,Jiang2021LightXMLTW}.
This paradigm has multiple real-world applications including movie and product recommendation, search-engines, and e-commerce product tagging.
In many of these applications, models are required to handle the addition of new classes on a regular basis, which has been the subject of recent work on zero-shot and few-shot extreme classification (\zsxc{} and \fsxc{})~\cite{Gupta2021GeneralizedZE,Xiong2022ExtremeZL,Simig2022OpenVE}.
% Historically, these zero-shot and few-shot extreme classification settings (\zsxc{} and \fsxc{}) have received little attention, with some recent works trying to tackle it~\citep{Gupta2021GeneralizedZE,Xiong2022ExtremeZL,Simig2022OpenVE}.
% To tackle this, recent works have developed models for 
These setups are challenging because of
(1) the presence of a large number of fine-grained classes which are often not mutually exclusive,
(2) limited or no labeled data per class, and
(3) increased computational expense and model size due to the large label space.
While the aforementioned works have tried to tackle the latter two issues, they lack a semantically rich representation of classes, and instead rely on class names or hierarchies to represent them.
% with ZestXML~\citep{Gupta2021GeneralizedZE}, MACLR~\citep{Xiong2022ExtremeZL}, LightXML~\citep{Jiang2021LightXMLTW}, and GROOV~\citep{Simig2022OpenVE} using only class names to represent them.

% A large fine-grained label space necessitates capturing the semantics of different attributes of classes.
In this work, we leverage semantic supervision (\semsup{})~\citep{Hanjie2022SemanticSE} for developing models for extreme classification. \semsupxc{} represents classes using multiple diverse class descriptions, which allows it to generalize naturally to novel classes when provided with corresponding descriptions.
However, \semsup{} as proposed in \cite{Hanjie2022SemanticSE} cannot be naively applied to XC for several reasons:
(1) \semsup{} requires encoding descriptions of all classes for each training batch, which is prohibitively computationally expensive for large label spaces,
% and we alleviate this in~\semsup{} by proposing a contrastive learning objective~\citep{hadsell2006dimensionality} which samples a fixed number of negative label descriptions, thus improving computation speed by \textcolor{red}{X\%} on~\eurlex{}.
(2) it uses semantic similarity only at the sentence level between the instance and label description, and
% , thus ignoring semantically/lexically similar words like ``\textit{photo}'' and ``\textit{picture}'' which indicate strong compatibility, and
% We remedy this by proposing a novel hybrid lexical-semantic similarity model called~\relaxed{} which is based on COIL~\citep{coil} and combines semantic similarity of sentences with soft matching between all token pairs.
(3) it requires human intervention to collect descriptions, which is expensive for extremely large label spaces we are dealing with.
%\todo{can we take a guess for how long it would take?}.

% We remedy these deficiencies by proposing~\semsupweb{} which is a fully automatic pipeline with precise heuristics to scrape high-quality descriptions.
% We remedy these deficiencies by developing a new model called \semsupxc{} which scales to large class spaces in XC using three innovations. First, \semsupxc{} employs a contrastive learning objective~\citep{hadsell2006dimensionality} which samples a fixed number of negative label descriptions, improving computation speed by as much as 99.9\%. Second, we use a novel hybrid lexical-semantic similarity model called~\relaxed{} (based on COIL~\citep{coil}) that combines semantic similarity of sentences with relaxed lexical-matching between all token pairs.
% Finally, we propose \semsupweb{} -- a fully automatic pipeline with precise heuristics to scrape high-quality descriptions.

We remedy these deficiencies by developing \semsupxc{}, a model that scales to large class spaces in XC and establishes a new state-of-the-art using three innovations.
First, we use a novel hybrid lexical-semantic similarity model (\relaxed{}) that combines semantic similarity of sentences with relaxed lexical-matching between all token pairs.
Second, we propose \semsupweb{} -- an automatic pipeline with precise heuristics to discover high-quality descriptions.
Finally, we use a contrastive learning objective
% ~\citep{hadsell2006dimensionality} 
that samples a fixed number of negative label descriptions, improving computation speed by up to 99\% when compared to~\semsup{}.

% improves \semsup{}'s training (through contrastive learning), modeling (through~\relaxed{}), and description quality (through~\relaxed{}) to
\semsupxc{} achieves state-of-the-art performance on three diverse XC datasets from legal (\eurlex{}), e-commerce (\amazon{}), and wiki (\wikipedia{}) domains, across zero-shot (\zsxc{}), generalized zero-shot (\gzsxc{}) and few-shot (\fsxc{}) settings.
For example, on~\zsxc{}, \semsupxc{} outperforms the next best baseline by $5-12$ precision points over all datasets and all metrics.
On \fsxc{}, \semsupxc{} consistently outperforms baselines by over $10$ \precision{1} points on the \eurlex{} and \amazon{} datasets.
Surprisingly, \semsupxc{} even outperforms larger unsupervised language models like T5 and Sentence Transformers (e.g., by over $30$ P@1 points on \eurlex{}) which are pre-trained on much larger web-scale corpora.
% , which shows the importance of contrastive learning to adapt to a specific domain.
Our ablation studies dissect the importance of each component in \semsupxc{}, and shows the importance of our proposed hybrid matching module.
Qualitative error analysis of our model (Table~\ref{table:qualitative}) shows that it predicts diverse correct classes that are applicable to the instance at times, whereas other models either predict incorrect classes or suffer a mode collapse.
% of the model shows its
% We find that \semsupxc{}'s hybrid matching module enables it to predict labels which are semantically or lexically similar.
% \karthik{can add one interesting finding from ablation, if any}

% \ameet{Need to write the ablation study, will do it after we have qualitative analysis results.}

%% file: LaTeX/Figure_Latex/teaser_figure.tex
%%%% Teaser figure %%%%
\begin{figure*}
\centering
\includegraphics[width=\linewidth]{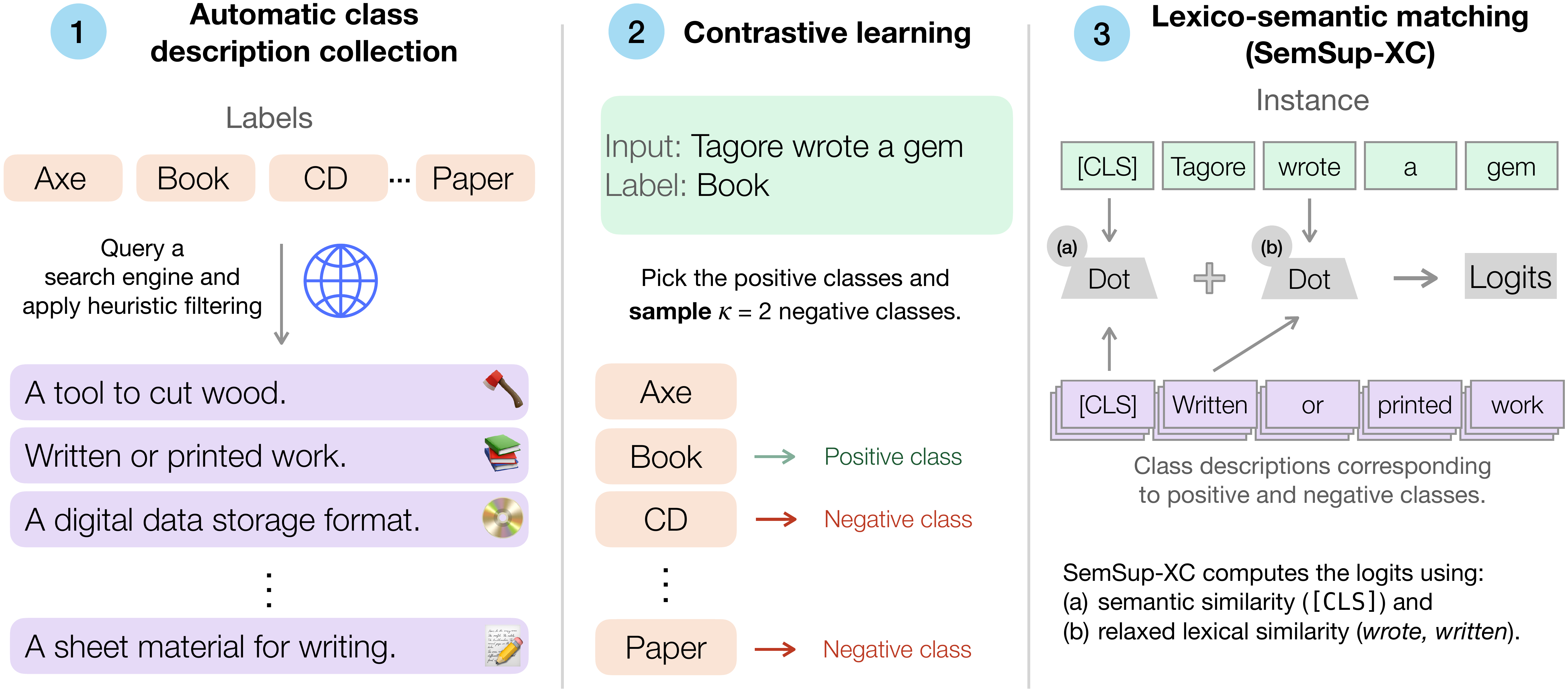}
\caption{Our model \semsupxc{} achieves state-of-the-art performance on zero-shot and few-shot extreme classification through three innovations -- \circled{1}{approach}{approach}{black} large-scale automated class description collection with heuristic filtering to improve semantic understanding of classes, \circled{2}{approach}{approach}{black} contrastive learning to make training faster by over 99\% when compared to the previous work (\semsup{}), and \circled{3}{approach}{approach}{black} a novel lexico-semantic matching model building called~\relaxed{} to utilize both semantic similarity at the sentence level and contextual lexico-semantic similarity at the token level.
% \circled{1}{approach}{approach}{black} and \circled{2}{approach}{approach}{black} significantly improve~\semsup{}'s computational speed (over 99\% on~\wikipedia{}), \circled{3}{approach}{approach}{black} boosts performance~\symbolsecref{sec:results:zero_shot}
% \ameet{Cite COIL in the Figure as well.}
% \ameet{Can we give a couple examples for heuristics?}
}
\label{fig:teaser}
\end{figure*}
%%%%%%%%%%%%%%%%%%%%%%%

%% file: LaTeX/400_methods.tex
\section{Methodology}
\label{sec:methodology}

% \ameet{Make sure I incorporate that negative sampling using TF-IDF scores. Cite this paper~\cite{zhang2022metadata}.}
% \ameet{Should I add a projection matrix between input and output encoder?}

%KN:  this is redundant text
% We briefly describe the semantic supervision framework (\semsup{})~\citep{Hanjie2022SemanticSE} in the context of multi-label classification, which entails problem where more than one label can be correct for each instance.
% Subsequently, we detail how our method (\semsupxc{}) improves \semsup{}'s task performance and computational complexity.

% \subsection{Problem Formulation: Zero and Few Shot Extreme Classification}
\subsection{Background}
\label{sec:methodology:background}
% \label{sec:methodology:semsup}
\paragraph{Extreme Classification}

% \note{Pranjal: This subsection is in response to Reviewer 4: Point 1 and Reviewer 2: Point 2. Currently it looks a bit redundant and can be shortened.}
% \pranjal{@Ameet, should I rather write this as 'notation section' and shorten it?}

Extreme classification deals with prediction over large label spaces (thousands to millions classes) and multiple correct classes per instance (multi-label)~\cite{Agrawal2013MultilabelLW,Bhatia2015SparseLE,Babbar2017DiSMECDS,Gupta2021GeneralizedZE,Xiong2022ExtremeZL,Simig2022OpenVE}.
Zero-shot extreme classification (\zsxc{}) is a variant where models are evaluated on unseen classes not encountered during training.
We evaluate both on (1) zero-shot (\zs{}), where the model is tested only on unseen classes and (2) generalized zero-shot (\gzs{}), where the model is tested on a combined set of train and unseen classes.
We also consider few-shot extreme classification (\fsxc{}), where a small number of supervised examples (e.g., $5$) are available for unseen classes.
% The large number of fine-grained classes with a heavy-tailed distribution in \xc{} pose efficiency and performance challenges.
The heavy tailed distribution of a large number of fine-grained classes in \xc{} poses efficiency and performance challenges.

% \vspace{-16pt}

\paragraph{Zero-shot classification}
% \label{sec:methodology:semsup}

Zero-shot classification is usually performed by matching instances to auxiliary information corresponding to classes, like their class name or attributes~\cite{larochelle2008zero,dauphin2014zero}.
Recently, \citet{Hanjie2022SemanticSE} proposed the use of \textit{multiple} class descriptions to endow the model with a holistic semantic understanding of the class from different viewpoints.
% \semsup{} uses an (1) input encoder (\inputencoder{}) to encode the instance and an (2) output encoder (\outputencoder{}) to encode class descriptions, and makes predictions by measuring the compatibility of the input and output representations.
\semsup{} uses an (1) input encoder (\inputencoder{}) to encode the instance and an (2) output encoder (\outputencoder{}) to encode class descriptions, and makes predictions by measuring the compatibility of the input and output representations.
Formally, let
% \numclasses{} be the number of classes,
% \hiddendim{} be the dimensionality of the input representation,
%KN: don't think we need to talk about supervised
% \outputmatrix{} $\in \mathbb{R}^{\textrm{\numclasses{}} \times \textrm{\hiddendim{}}}$ be the randomly initialized output matrix used by \supervised{} to classify the input, with the $j^{\textrm{th}}$ row corresponding to the $j^{\textrm{th}}$ class,
$x_i$ be the input instance,
% $\mathcal{D} = \left ( d_1, \dots, d_{\textrm{\numclasses{}}} \right )$ be descriptions corresponding to the classes,
$d_j \in \mathcal{D}_j$ be one sampled description of class $j$,
$\textrm{\inputencoder}\left ( x_i \right ), \textrm{\outputencoder}\left ( d_j \right )\in \mathbb{R}^{d}$ be the input and output representation respectively.
% and $\textrm{\outputencoder}\left ( d_j \right )$ $\in \mathbb{R}^{\textrm{\hiddendim{}}}$ be the output representation of the $j^{\textrm{th}}$ class.
%  be a matrix such that the $i^{\textrm{th}}$ row be the output representation of the $i^{\textrm{th}}$ class.
For the multi-label XC setting, the 
% For multi-label classification, the following mathematically defines the 
probability of picking the $j^{\textrm{th}}$ class is:
\begin{align}
    \label{eq:semsup_loss}
    \begin{split}
        % \textrm{\supervised{}}& \coloneqq P(y_j = 1|x_i) = \sigmoid \left ( \mathcal{O}[j] \cdot \textrm{\inputencoder}\left ( x_i \right )   \right ) \\
        \textrm{\semsup{}}& \coloneqq P(y_j = 1|x_i) = \sigma \left ( \textrm{\outputencoder}\left ( d_j \right )^\intercal \cdot \textrm{\inputencoder}\left ( x_i \right )  \right )
    \end{split}
\end{align}

\paragraph{Challenges with large class spaces}
\citet{Hanjie2022SemanticSE}'s method cannot be directly applied to \xc{} for several reasons.
First, they use a bi-encoder model~\cite{bai2009supervised} which measures only the semantic similarity between the input instance and the class description at the sentence level.
However, instances and descriptions often share lexical \textit{terms} with the same or similar meaning and lemma (e.g., \textit{picture} and \textit{photo}), which is not exploited by their method.
Second, their class description collection pipeline requires human intervention, which is not feasible for the large number of classes in \xc{} datasets.
And third, they fine-tune the label encoder by encoding descriptions for \textit{all} classes for every batch of instances, making it computationally infeasible for large label spaces because of GPU memory constraints.

\input{LaTeX/401_methods_semsup_xc.tex}

%% file: LaTeX/401_methods_semsup_xc.tex
\subsection{\semsupxc{}: Improved \zs{} extreme classification}
\label{sec:methodology:semsupxc}

\semsupxc{} addresses the aforementioned challenges using:
(1) a novel hybrid semantic-lexical similarity model for improved performance,
(2) an automatic class description discovery pipeline with accurate heuristics for garnering high-quality class descriptions,
and
(3) contrastive learning with negative samples for improved computational speed
.

\subsubsection{Hybrid lexico-semantic similarity model}

\semsup{}'s bi-encoder architecture measures only the semantic similarity of the input instance and class description at sentence level.
% Drawing inspiration from recent IR models like~\coil{}~\citep{coil} and ColBERT~\citep{colbert}, we note that 
However, semantic similarity ignores lexical matching of shared words which exhibit strong evidence of compatibility. (eg., Input: \textit{It was \underline{cold} and flavorful} and Description: \textit{Ice cream is a \underline{cold} dessert}).
% \todo{Pranjal: Add an example here.}
% \todo{Pranjal: The athletics example is not very clear. Give short examples of instances and labels.}
% , thus leading to performance degradation.
A recently proposed information-retrieval model, \coil{}~\citep{coil}, alleviates this by incorporating lexical similarity by adding the dot product of contextual representations corresponding to common tokens between the query and document.
% However, sentences can have tokens which are \textit{similar} but not the \text{same}, and \coil{} ignores such token pairs.
But \coil{} has the drawback that semantically similar tokens (e.g., ``pictures'' and ``photos'') and words with the same lemma (e.g., walk and walking) are treated as dissimilar tokens, despite being commonly used interchangeably. (eg., Input: \textit{Capture the best moments in high quality \underline{pictures}} and Class description: \textit{A camera is used to take \underline{photos}}.)
% For example, the two underlined tokens in the following sentences: ``Capture the best moments in high quality \underline{pictures}'' and ``A camera is used to take \underline{photos}.''

% \ameet{In the following section, we should refer to the appendix twice. Once after the first line, because people will be confused how the clusters are created, and once towards the end, just mentioning that the details are in ...
% Might also help if we provide an example.
% }
We propose \relaxed{} to exploit such token similarity.
%  between such token pairs, we propose a model called \relaxed{} which incorporates relaxed lexical-matching.
%  and considers \textit{similar} tokens.
% We create clusters of tokens based on: 1) the BERT token-embedding similarity~\citep{Rajaee2021ACAisotropic}  and 2) the lemma of the word, where two tokens are in same cluster if either embedding similarity is higher than a threshold or both share the same lemma. We consider tokens belonging to the same cluster to be similar.
We create clusters of tokens based on: 1) the BERT \textit{token-embedding similarity}~\citep{Rajaee2021ACAisotropic} is higher than a threshold or 2) if tokens share the same \textit{lemma}, which results in tokens like ``photo'', ``picture'', and ``pictures'' being in the same cluster.  
We provide implementation details on clustering in Appendix~\ref{app:hybridImplementation}.
% \relaxed{} incorporates semantic similarity by taking the dot-product of \texttt{[CLS]} representations of the input instance and the class description.
In addition to semantic similarity, \relaxed{} uses these clusters to for relaxed lexical-matching by computing the dot-product of contextual representations of ``similar'' tokens in the input and description, as judged by the clusters.
For cases where an input token has several similar tokens in the description, we choose the description token with the max dot product with the former.
% The scores are aggregated for all the tokens in the input, and if an input token does not have a similar token in the description, it is ignored.
% \coil{}~\citep{coil} is a bi-encoder architecture which alleviates this by incorporating both semantic and lexical similarity.
% Apart from the dot product between \clstoken{} vectors, they also include an exact lexical match scoring function which is based on the dot product of representations corresponding to tokens with exact matches in the two pieces of text considered (e.g., input text: 
% ``Capture the best momemts in high quality \underline{pictures}.'' and label description ``A camera is used to take \underline{photos}.''
% % ``My kitty used her \underline{paws} to play with the ball'' and label description ``A cat is an animal with nimble \underline{paws}'').
% % \todo{UPDATED \sout{Can we add a sensible example from one of the extreme classification datasets above? We need to use the same one in the figure}} \karthik{+1}
% If there are multiple occurrences of a common token type, the maximum similarity score is chosen, and the scores are then aggregated over all token types that are present in both sentences.
Formally let $x_i = (x_{i1}, \dots, x_{in})$ be the input instance with $n$ tokens,
$d_j = (d_{j1}, \dots, d_{jm})$ class $j^{\textrm{th}}$ descriptions with $m$ tokens,
$v_{\textrm{cls}}^{x_i}$ and $v_{\textrm{cls}}^{d_j}$ be the \clstoken{} representations of the input and description,
and $v_{k}^{x_i}$ and $v_{l}^{d_j}$ be the representation of the $k^{\textrm{th}}$ and $l^{\textrm{th}}$ token of $x_i$ and $d_j$ respectively.
Let $\textrm{CL}(w)$ denote the cluster membership of the token $w$, with $\textrm{CL}(w_i) = \textrm{CL}(w_j)$ implying that the tokens are \textit{similar}.
Then probability of class $y_j$ is:
% \relaxed{} computes the probability as follows:
\begin{align}
    \begin{split}
        & \textrm{\relaxed{}} \coloneqq P(y_j = 1|x_i) = \sigma \biggl( {v_{\textrm{cls}}^{x_i}}^\intercal \cdot v_{\textrm{cls}}^{d_j} \\ & + \sum_{k=1}^n \max_{l\in\{1,\dots,m\},~\textrm{CL}\left ( x_{ik}\right ) = \textrm{CL} \left ( d_{jl} \right ) } \left ( {v_{k}^{x_i}}^\intercal v_{l}^{d_j} \right ) \biggr)
    \end{split} 
\end{align}

% \karthik{idx = CL(x) = CL(d) above is unclear. how does idx come from intersection of $x_i$ and $d_j$?}

% \input{LaTeX/Tables/zsl_table3.tex}

\subsubsection{\semsupweb{}: Automatic collection of high-quality descriptions}
% \semsup{} uses a semi-automatic pipeline for collecting multiple descriptions for classes, with an expert required for filtering irrelevant ones.
% However, in our case, large label spaces (e.g., 1 million for \wikipedia{}) make any degree of human involvment infeasible.
We create a completely automatic pipeline for collecting descriptions which includes sub-routines for removing spam, advertisements, and irrelevant descriptions, and we detail the list of heuristics used in Appendix~\ref{app:descriptions}.
These sub-routines contain precise rules to remove irrelevant descriptions, for example by removing sentences with too many special characters (usually spam), descriptions with clickbait phrases (usually advertisements), ones with multiple interrogative phrases (usually people's comments), small descriptions (usually titles) and so on (Appendix~\ref{app:descriptions}). Further in \wikipedia{}, querying search engine for labels return no useful results, since labels are very specific (eg., Fencers at the 1984 Summer Olympics) . Therefore, we design a multi-stage approach, where we first break label names into relevant constituents and query each of them individually (see Appendix~\ref{app:desc_wiki}). 
In addition to web-scraped label descriptions, we utilize label-hierarchy information if provided by the dataset (\eurlex{} and \amazon{}), which allows us to encode properties about parent and children classes wherever present.
Further details for hierarchy are present in Appendix~\ref{app:desc_hier}.
As we show in the ablation study~\symbolsecref{sec:results:ablations}, label descriptions that we collect automatically provide significant performance boosts.

%%%%%%%%%%%%%%%% Old %%%%%%%%%%%%%%%%

\subsubsection{Training using contrastive learning}
For datasets with a large number of classes (large $|C|$), it is not computationally feasible to encode class descriptions for all classes for every batch.
We draw inspiration from contrastive learning~\citep{hadsell2006dimensionality} and sample a significantly smaller number of negative classes to train the model.
% For datasets with a large label space (large $|C|$), we improve \semsup{}'s computational speed by sampling negative classes for each instance rather than encoding the label descriptions of all classes.
For an instance $x_i$, consider two partitions of the labels $Y_i=\{y_{i1}, \dots, y_{iC}|y_{ij} \in \{0, 1\} \}$, with $Y_i^{+}$ containing the positive classes ($y_{ij}=1$) and $Y_i^{-}$ containing the negative classes ($y_{ij}=0$).
\semsupxc{} caps the total number of class descriptions being encoded for this instance to $\mathcal{K}$ by using all the positive classes ($\vert Y_i^{+} \vert$) and sampling only $\mathcal{K} - \vert Y_i^{+} \vert$ negative classes.
%  from $Y_i^{-}$.
% \semsupxc{} is trained using all positive classes ($\vert Y_i^{+} \vert$), but drawing inspiration from contrastive learning~\citep{hadsell2006dimensionality}, we sample $K - \vert Y_i^{+} \vert$ negative classes from $Y_i^{-}$ instead of $C - \vert Y_i^{+} \vert$, with $K \ll \vert C \vert$.
Intuitively, our training objective incentivizes the representations of the instance and positive classes to be similar while simultaneously making them dissimilar to the negative classes.
To improve learning, rather than picking negative labels at random, we sample hard negatives that are lexically similar to positive classes.
A typical dataset we consider (\amazon{}) has $\vert C \vert = 13,000$ and $\mathcal{K}$ $\approx 1000$, which leads to \semsupxc{} being $\frac{12000}{13000}=92.3\%$ faster than~\semsup{}.
Mathematically, the following is the training objective, where $N$ is the train dataset size.
{
\begin{align}
    \begin{split}
        \mathcal{L}_{\textrm{\semsupxc{}}} = \dfrac{1}{N\cdot \mathcal{K}} & \sum_i \biggl( \sum_{y_k \in Y_i^+} \mathcal{L}_{\textrm{BCE}}\left ( P \left (y_k = 1 | x_i \right ), y_k \right )  \\ + & \sum_{y_l \sim Y_i^-,~l=1}^{\mathcal{K} - \vert Y_i^{+} \vert} \mathcal{L}_{\textrm{BCE}}\left ( P(y_l = 0 | x_i), y_l \right )  \biggr) \label{eq:semsupxc_loss}
    \end{split}
\end{align}
}
For each batch, a class description is randomly sampled for each class ($d_j^l \in \mathcal{D}_j$), thus allowing the model to see all the descriptions in $\mathcal{D}_j$ over the course of training, with the same sampling strategy during evaluation. 
We refer readers to appendix~\ref{app:contrastive} for additional details.

%% file: LaTeX/500_experimental_setup.tex
\input{LaTeX/Tables/dataset_table.tex}
\input{LaTeX/Tables/zsl_table3.tex}

\section{Experimental Setup}
\label{sec:experimentalsetup}

\paragraph{Datasets}
We evaluate our model on three diverse public datasets.
They are, \textbf{\eurlexfull{}}~\citep{Chalkidis2019LargeScaleMT} which is legal document classification dataset with 4.3K classes,
% In the test set, the official dataset comprises only 163 unseen labels in the zero-shot setting. To make the test set more competitive, we randomly sample from the test labels, and build a more challenging split with 1057 unseen labels.
\textbf{\amazonfull{}}~\citep{McAuley2013HiddenFA} which is an e-commerce product tagging dataset including Amazon product descriptions and titles with 13K categories,
% Because there are no unseen labels in the dataset, we generate a new test split with 6500 labels by randomly sampling from all of them.
and \textbf{\wikipediafull{}}~\citep{Gupta2021GeneralizedZE} which is an article classification dataset made up of 5 million Wikipedia articles with over 1 million categories.
We provide detailed statistics about the number of instances and classes in train and test set in Table~\ref{table:dataset-stats}. See Appendix~\ref{app:splitcreation} for details on split creation. 

%%%%%%%%%% Old %%%%%%%%%%
% We train and evaluate on the same splits used in the original paper.
% \begin{itemize}
%     \item \textbf{EURLex-4.3K:} is a dataset of 4.3K labels for legal document classification.
%     In the test set, the official dataset comprises 163 unseen labels, with just 89 input documents in the zero-shot mode.
%     To make the test set more competitive, by randomly sampling from the test labels, we build a more challenging zero-shot split with 1057 unseen labels.
%     We provide results for both of these splits.
%     \item \textbf{AmazonCat-13K:} includes Amazon product descriptions and titles, with the task of categorizing them into one of 13K categories.
%     Because there are no unseen labels in the dataset, we generate a new test split with 6500 labels by randomly sampling from all of them.
%     \item \textbf{Wiki-1M:} is made up of over 5 million Wikipedia articles and 1 million categories.
%     We train and evaluate on the same splits that were used in the original paper.
% \end{itemize}
%%%%%%%%%% Old %%%%%%%%%%

\paragraph{Baselines}
% \ameet{Let's add back the unsupervised and supervised distinction you had before. My bad.}
We perform extensive experiments with several baselines, which can be divided into \textit{unsupervised} (first three) and \textit{supervised} which are fine-tuned on the datasets we consider (the remaining four).
% The first three of the following baselines are unsupervised, and the remaining four are supervised.
%  which can be divided into unsupervised and supervised Methods.
% Unsupervised methods are not trained on any of our datasets, and are directly evaluated on the dataset.
\textbf{1)~TF-IDF} performs a nearest neighbour match between the sparse tf-idf features of the input and class description.
\textbf{2)~T5}~\citep{t5_google} is a large sequence-to-sequence model which has been pre-trained on 750GB unsupervised data and further fine-tuned on MNLI~\citep{MNLI}. We evaluate the model as an NLI task where labels are ranked based on the likelihood of entailment to the input document. For computational efficiency, we evaluate T5 only on top 50 labels shortlisted by \tfidf{} on each instance.
% Ranking is done on top 50 labels predicted by \tfidf{}
\textbf{3)~Sentence Transformer}~\citep{Reimers2019SentenceBERTSE} is a semantic text similarity model fine-tuned using a contrastive learning objective on over 1 billion sentence pairs. 
We rank the labels based on the similarity between input and description embeddings.
The latter two baselines use significantly more data than~\semsupxc{} and T5 has $9\times$ the parameters.
The aforementioned baselines are unsupervised and not fine-tuned on our datasets. 
The following baselines are previously proposed supervised models and are fine-tuned on the datasets we consider.
\textbf{4)~ZestXML}~\citep{Gupta2021GeneralizedZE} learns a highly sparsified linear transformation ( W ) which projects sparse input features close to corresponding positive label features. At inference, for each input instance $x_i$, label $l_j$ is scored based on the formula $s_{ij} = l_j^T  W x_{i}$.
\textbf{5)~MACLR}~\citep{Xiong2022ExtremeZL} is a bi-encoder based model pre-trained on two self-supervised learning tasks to improve extreme classification---Inverse Cloze Task~\citep{InverseCloze} and SimCSE~\citep{Gao2021SimCSESC}, and we fine-tune it on the datasets considered.
% For a fair comparison, we further fine-tune the model on complete train set. (\todo{UPDATED: mention}) ...
\textbf{6)~GROOV}~\citep{Simig2022OpenVE} is a T5 model that learns to generate labels given an input instance.
\textbf{7)~SPLADE}~\citep{Formal2021SPLADESL} is a state-of-the-art sparse neural retreival model that learns label/document sparse expansion via a Bert masked language modelling head.
% It is one of the current state-of-the-art models in information retrieval for out-of-domain tasks.

% We evaluate two variants of the baselines, one that uses class names as the auxiliary information and another which uses the same class descriptions used by our \semsupxc{} models.
% We use the additional label hierarchy information in all cases.
% The version of baselines which use our class descriptions are strictly comparable to \semsupxc{} models.
% \todo{Pranjal: You can make the previous part shorter.}

We evaluate baselines under two different settings -- by providing either class names or class descriptions as auxillary information, and use the label hierarchy in both settings.
The version of baselines which use our class descriptions are strictly comparable to \semsupxc{} models.
See Appendix~\ref{app:baselines} for additional details.

% We fine-tune all the supervised models and our~\semsupxc{} models on the datasets and use the same class descriptions to make comparisons fair.
% \todo{}
%  with~\semsupxc{} fair, we fine-tune and re-evaluate the above models on the datasets we consider while including label descriptions and label hierarchy information. 
% We refer readers to Appendix~\ref{app:baselines} for additional details.

% \pranjal{UPDATED: I think we are using LightXML now? You can change the following.}
% \textbf{7) LightXML}~\citep{Jiang2021LightXMLTW} is state of art extreme classifier, which uses a transformer backbone and samples hard negatives using a generative model.
% We defer implementations details of the model to Appendix~\todo{reference the correct appendix section here.}

\paragraph{\semsupxc{} implementation details}
We use the Bert-base model~\citep{Devlin2019BERTPO} as the backbone for the input encoder and Bert-small model~\citep{Turc2019WellReadSL} for the output encoder.
\semsupxc{} follows the model architecture described in Section~\ref{sec:methodology:semsupxc} (\relaxed{}) and we use contrastive learning~\cite{hadsell2006dimensionality} to train our models.
During training, we sample $\mathcal{K} - |Y_i^+|$ hard negatives for each instance, where $\mathcal{K}$ is the number of labels for the instance.
% \pranjal{It is not 1000 right? I though it is 1000 - $p$? And can we explain the above a little better?}
At inference, to improve computational efficiency, we precompute the output representations of label descriptions and shortlist top 1000 labels based on the \tfidf{} scores. 
% Similar to \zest{}, we sum over logits from \semsupxc{} and \tfidf{}.
We use the AdamW optimizer~\citep{Loshchilov2019DecoupledWD} and tune our hyperparameters using grid search on the respective validation set. We use similar hyperparameters for all datasets, and similar settings across all baselines.
See Appendix~\ref{app:hyperparameter} for more details.
% \todo{Appendix and detail about few-shot fine-tuning}

% \pranjal{We should probably also include a detaul about few-shot classification fine-tuning}

% \vspace{-0.3cm}
\paragraph{Evaluation setting and metrics}
We evaluate all models on three different settings:
Zero-shot classification (\zsxc{}) on a set of unseen classes,
generalized zero-shot classification (\gzsxc{}) on a combined set of seen and unseen classes,
and few-shot classification (\fsxc{}) on a set of classes with minimal amounts of supervised data ($1$ to $20$ examples per class).
For all three settings, we train on input instances of seen classes.
We use Precision@K and Recall@K as our evaluation metrics, as is standard practice.
Precision@K measures how accurate the top-K predictions of the model are, and Recall@K measures what fraction of correct labels are present in the top-K predictions, and they are mathematically defined as $P@k=\frac{1}{k} \sum_{i \in \operatorname{rank}_k(\hat{y})} y_i$ and $R@k=\frac{1}{\sum_{i} y_i} \sum_{i \in \operatorname{rank}_k(\hat{y})} y_i$, where $\operatorname{rank}_k(\hat{y})$ is the set of top-K predictions. We average the metrics over test instances.
%  $\hat{y}$ is the prediction vector and y is the ground truth vector.
\input{LaTeX/Result_files/015_few_shot_figure}

% \pranjal{What do we do with multiple descriptions? We should probably descibe that here.}

% \begin{equation}
% P@k=\frac{1}{k} \sum_{i \in \operatorname{rank}_k(\hat{y})} y_i ,
% \end{equation}
% \begin{equation}
% R@k=\frac{1}{\sum_{i} y_i} \sum_{i \in \operatorname{rank}_k(\hat{y})} y_i ,
% \end{equation}
% where $\hat{y}$ is the prediction vector and y is the ground truth vector.

% \subsubsection{Unsupervised Methods}
% \note{These methods are not trained on any of the XMC datasets.} 
% \begin{enumerate}
%     \item \textbf{TF-IDF:} Nearest Neighbour matching on the sparse TF-IDF features between the input document and label names.
%     \item \textbf{T5-Large:} a sequence-to-sequence encoder-decoder based transformer model, pretrained on mnli and xtreme-xnli datasets.
%     \item \textbf{Sentence Transformer:} a sentence encoder model, trained using a contrastive learning objective on 1B sentence/paragraph pairs.
% \end{enumerate}

% \subsubsection{Supervised Methods}
% \begin{enumerate}
%     \item \textbf{ZestXML:} learns a sparsified linear transformation between input features and the label features. We use the code provided by the authors for training and inference. 
%     \item \textbf{MACLR:} uses several pretraining tasks to perform unsupervised classification on the target split. However to make comparisons more fair, we completely finetune the model on the train split.
%     \item \textbf{GROOV:} a sequence to sequence tranformer model that learns to generate labels. At inference it generates seen as well as novel labels.
%     \item \textbf{AttentionXML:}
    
% \end{enumerate}

%% file: LaTeX/Tables/dataset_table.tex
\begin{table}[t]
\centering
\resizebox{\dimexpr\columnwidth  }{!}{%
\begin{tabular}{cccccc}
\toprule
\multirow{2}{*}{\bf{Dataset}} & 
\multicolumn{3}{c}{\bf{Documents}} & 
\multicolumn{2}{c}{\bf{Labels}} \\
\cmidrule(lr){2-4} \cmidrule(lr){5-6}
& $\bf{N_{\text {train }}}$ & $\bf{N_{\text {test }}}$ &  $\bf{N_{\text {test(\zsxc)}}}$ & $\bf{\left|Y_{\text {seen }}\right|}$ & $\bf{\left|Y_{\text {unseen }}\right|}$ \\
\midrule
% \hline \hline 
EURLex-4.3K & $45 \mathrm{~K}$ & $6 \mathrm{~K}$ & $5.3 \mathrm{~K}$ & 3,136 & 1,057 \\
AmazonCat-13K & $1.1 \mathrm{M}$ & $307 \mathrm{K}$ & $268 \mathrm{~K}$ & 6,830 & 6,500 \\
Wikipedia-1M & $2.3 \mathrm{M}$ & $2.7 \mathrm{M}$ & $2.2 \mathrm{M}$ & 495,107 & 776,612 \\
\hline
\end{tabular}}
\caption{
Dataset statistics along with information about zero-shot (\zsxc) splits. $N_{\textrm{testzsl}}$ indicates number of samples in zero-shot split, and $Y_{avg}$ indicates average number of positive labels per input document.
} 
\label{table:dataset-stats}
\end{table}

% Table 2: Extreme Zero-shot Learning (EZ-XMC) comparison of different unsupervised methods.
% \begin{table}[t]
% \centering
% \resizebox{\dimexpr\columnwidth  }{!}{%
% \begin{tabular}{ccccccc}
% \toprule
% \multirow{2}{*}{\bf{Dataset}} & 
% \multicolumn{4}{c}{\bf{Documents}} & 
% \multicolumn{2}{c}{\bf{Labels}} \\
% \cmidrule(lr){2-5} \cmidrule(lr){6-7}
% & $\bf{N_{\text {train }}}$ & $\bf{N_{\text {test }}}$ &  $\bf{N_{\text {testzsl}}}$ & $\bf{|Y_{avg}|}$ & $\bf{\left|Y_{\text {seen }}\right|}$ & $\bf{\left|Y_{\text {unseen }}\right|}$ \\
% \midrule
% % \hline \hline 
% EURLex-4.3K & $45 \mathrm{~K}$ & $6 \mathrm{~K}$ & $5.3 \mathrm{~K}$ & 547.5 & 3,136 & 1,057 \\
% AmazonCat-13K & $1.1 \mathrm{M}$ & $307 \mathrm{K}$ & $268 \mathrm{~K}$ & $210.4$ & 6,830 & 6,500 \\
% Wikipedia-1M & $2.3 \mathrm{M}$ & $2.7 \mathrm{M}$ & $2.2 \mathrm{M}$ & $275.6$ & 495,107 & 776,612 \\
% \hline
% \end{tabular}}
% \caption{
% Dataset statistics along with information about zero-shot (\zsxc) splits. $N_{\textrm{testzsl}}$ indicates number of samples in zero-shot split%, and $Y_{avg}$ indicates average number of positive labels per input document.
% } 
% \label{table:dataset-stats}
% \end{table}

%% file: LaTeX/Tables/zsl_table3.tex
% Table 2: Extreme Zero-shot Learning (EZ-XMC) comparison of different unsupervised methods.
\begin{table*}[!t]
\centering
\resizebox{\linewidth}{!}{%
\begin{tabular}{lccccccccccccccc}
\toprule
\multirow{3}{*}{\bf{Model}} & \multicolumn{4}{c}{\textbf{\eurlexfull{}}} & \multicolumn{4}{c}{\bf{\amazonfull{}}} & \multicolumn{4}{c}{\bf{\wikipediafull{}}} \\
\cmidrule(lr){2-5} \cmidrule(lr){6-9} \cmidrule(lr){10-13}

&
\multicolumn{2}{c}{\bf{\zsxc{}}} & \multicolumn{2}{c}{{\bf{\gzsxc{}}}} &
\multicolumn{2}{c}{\bf{\zsxc{}}} & \multicolumn{2}{c}{{\bf{\gzsxc{}}}} &
\multicolumn{2}{c}{\bf{\zsxc{}}} & \multicolumn{2}{c}{{\bf{\gzsxc{}}}}
\\
\cmidrule(lr){2-3} \cmidrule(lr){4-5} \cmidrule(lr){6-7} \cmidrule(lr){8-9} \cmidrule(lr){10-11} \cmidrule(lr){12-13}

&
\bf{\precision{1}} & \bf{\recall{10}} & \bf{\precision{1}} & \bf{\recall{10}} & \bf{\precision{1}}  & \bf{\recall{10}} & \bf{\precision{1}} & \bf{\recall{10}} & \bf{\precision{1}} & \bf{\recall{10}} & \bf{\precision{1}} & \bf{\recall{10}}   \\
\midrule

\rowcolor{gray!20} & \multicolumn{12}{c}{\bf{Baselines with Class Names}} \\ \midrule
% \rowcolor{gray!20} \multicolumn{13}{c}{\bf{Baselines with Class Names}} \\ \midrule
\vspace{0.16cm}
\textit{\textbf{Unsupervised Baselines}} &\multicolumn{12}{c}{} \\
TF-IDF & $44.0$ & $55.8$ & $53.4$ & $41.2$ & $18.7$ & $21.0$ & $21.5$ & $14.7$ & $14.5$ & $18.3$ & $14.4$ & $14.7$\\
T5~\citep{t5_google} & $7.2$ & $29.2$ & $10.4$ & $23.0$ & $2.5$ & $10.5$ & $3.2$ & $10.2$ & $8.2$ & $23.6$ & $4.2$ & $15.1$\\
Sent. Transformer~\citep{Reimers2019SentenceBERTSE} & $16.6$ & $23.2$ & $20.9$ & $42.0$ & $18.2$ & $25.0$ & $21.1$ & $17.9$ & $7.8$ & $13.3$ & $5.2$ & $9.1$\\
\midrule
\vspace{0.16cm}
\textit{\textbf{Supervised Baselines}} &\multicolumn{12}{c}{} \\
% ZestXML~\citep{Gupta2021GeneralizedZE} & $9.6$ & $25.7$ & $84.8$ & $54.8$ & $12.7$ & $21.2$ & $87.9$ & $52.5$ & $12.9$ & $20.0$ & $26.7$ & $25.7$\\
% ZestXML + TF-IDF & $24.7$ & $46.4$ & $84.9$ & $60.2$ &  $15.6$ & $24.4$ & $87.6$ & $54.2$ & $15.8$ & $20.8$ & $26.3$ & $17.2$\\
ZestXML~\citep{Gupta2021GeneralizedZE} & $24.7$ & $46.4$ & $84.9$ & $60.2$ &  $15.6$ & $24.4$ & $87.6$ & $54.2$ & $15.8$ & $20.8$ & $26.3$ & $17.2$\\

% LightXML} & xx.y & xx.y & xx.y & xx.y & xx.y & xx.y & xx.y & xx.y & xx.y & xx.y & xx.y & xx.y\\
SPLADE~\citep{Formal2021SPLADESL} & $20.2$ & $24.4$ &  $52.3$ & $34.2$ & $17.2$  & $28.7$ & $ 75.8$ &  $41.3$  & $14.3$ & $17.8$ & $20.3$ & $22.4$\\
MACLR~\citep{Xiong2022ExtremeZL} & $24.9$ & $42.1$ & $60.7$ & $55.2$ & $36.0$ &  $54.4$ & $46.0$  & $46.9$ & $29.8$ & $41.7$ & $28.0$ & $32.7$\\
GROOV~\citep{Simig2022OpenVE} & $1.2$ & $7.0$ & $84.1$ & $49.4$ & $0.0$ & $2.4$ & $87.4$ & $47.9$ & $6.0$ & $15.4$ & $31.4$ & $29.0$\\

\midrule

% \bf{Auxiliary information: \semsupxc{} scraped class descriptions} & & & & & & & & & & &  \\ \cdashline{1-13}
\rowcolor{gray!20} & \multicolumn{12}{c}{\bf{Baselines with \semsupxc{} scraped Class Descriptions}} \\ \midrule
\vspace{0.16cm}
\textbf{\textit{Unsupervised Baselines}} &\multicolumn{12}{c}{} \\
TF-IDF & $43.7$ & $50.4$ & $57.2$ & $39.5$ & $17.4$ & $20.8$ & $21.1$ & $15.0$ & $9.2$ & $12.5$ & $9.1$ & $10.3$\\
T5~\citep{t5_google} & $5.0$ & $24.8$ & $3.3$ & $8.1$ & $2.8$ & $7.7$ &  $3.2$ & $4.2$  & $3.7$ & $13.4$ & $3.4$ & $13.2$\\
Sent. Transformer~\citep{Reimers2019SentenceBERTSE} & $15.9$ & $31.1$ & $18.8$ & $25.5$ & $15.2$ & $22.2$ & $16.0$ & $18.4$ & $19.6$ & $22.5$ & $14.2$ & $16.6$\\

\midrule
\vspace{0.16cm}
% \hl{\textbf{\textit{Supervised}}} &\multicolumn{12}{c}{} \\
\textbf{\textit{Supervised Baselines}} &\multicolumn{12}{c}{} \\

% ZestXML~\citep{Gupta2021GeneralizedZE} & $9.7$ & $25.8$ & $83.8$ & $55.2$ & $3.8$ & $20.5$ & $71.1$ & $49.7$ & $9.1$ & $13.9$ & $19.2$ & $17.8$\\
% ZestXML + TF-IDF & $22.6$ & $44.6$ & $84.2$ & $60.7$ &  $5.4$ & $24.8$ & $76.9$ & $50.7$ & $10.6$ & $14.1$ & $20.9$ & $17.9$\\
ZestXML~\citep{Gupta2021GeneralizedZE} & $22.6$ & $44.6$ & $84.2$ & $60.7$ &  $5.4$ & $24.8$ & $76.9$ & $50.7$ & $10.6$ & $14.1$ & $20.9$ & $17.9$\\

% \bf{LightXML} & xx.y & xx.y & xx.y & xx.y & xx.y & xx.y & xx.y & xx.y & xx.y & xx.y & xx.y & xx.y\\
SPLADE~\citep{Formal2021SPLADESL} & $20.7$ & $22.0$ & $45.1$ & $32.9$ & $16.9$  & $28.9$ & $77.0$ &  $42.0$  & $8.2$ & $11.1$ & $20.7$ & $22.4$ \\
MACLR~\citep{Xiong2022ExtremeZL} & $20.9$ & $37.9$ & $60.3$ & $53.8$ & $18.4$ &  $22.3$ & $36.5$  & $23.8$ & $30.7$ & $\mathbf{41.9}$ & $28.1$ & $33.6$ \\
GROOV~\citep{Simig2022OpenVE} & $0.3$ & $0.6$ & $80.2$ & $18.1$ & $0.0$ & $0.0$ & $84.5$ & $23.5$ & $0.5$ & $0.2$ & $7.0$ & $1.5$\\

\midrule
\rowcolor{gray!20} & \multicolumn{12}{c}{\bf{\semsupxcbig{} (Our Model)}} \\ \midrule

 \bf{\semsupxcbig{}} & $\mathbf{49.3}$ & $\mathbf{62.4}$ & $\mathbf{87.0}$ & $\mathbf{62.9}$ & $\mathbf{48.2}$ & $\mathbf{72.9}$ & $\mathbf{88.6}$ & $\mathbf{71.6}$ & $\mathbf{36.5}$ & $38.5$ & $\mathbf{33.7}$ & $\mathbf{34.1}$\\

\bottomrule

\end{tabular}
% Resize
}
\caption{
Zero-shot (\zsxc) and generalized zero-shot (\gzsxc) results for all models on three XC benchmarks. \semsupxc{} significantly outperforms state-of-the-art models on both precision (P@) and recall (R@) metrics across the board.}
\label{table:results-zsl3}
\end{table*}

%% file: LaTeX/Result_files/015_few_shot_figure.tex
\begin{figure*}[!htb]
% \ificlr
\centering
% \fi
\begin{subfigure}{.495\linewidth}
  \centering
  \includegraphics[width=1.0\columnwidth]{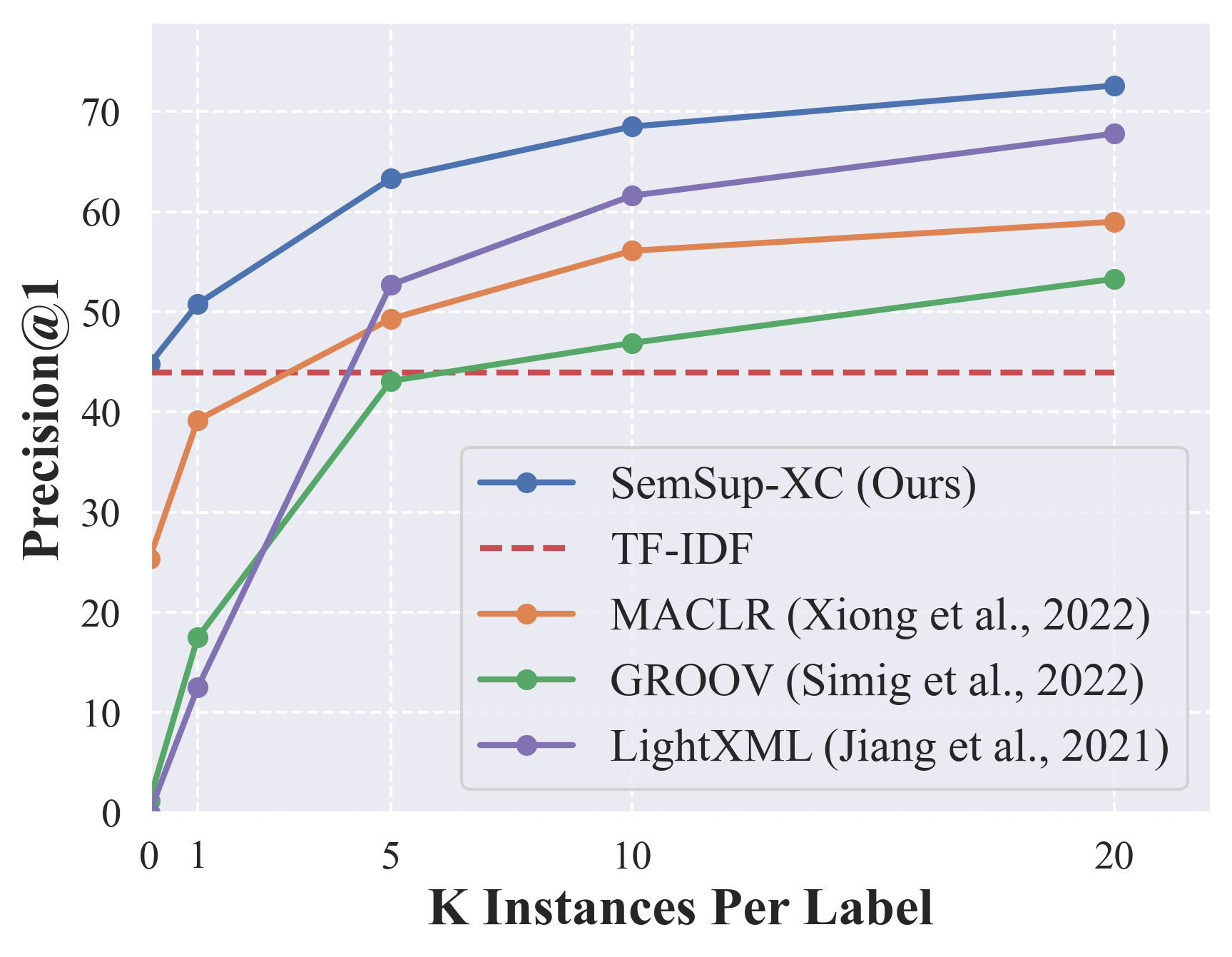}
  \caption{\eurlexfull{}}
  \label{fig:fsp1_eurlex}
\end{subfigure}%
 \hspace{0.01\columnwidth}
\begin{subfigure}{.495\linewidth}
  \centering
  \includegraphics[width=1.0\columnwidth]{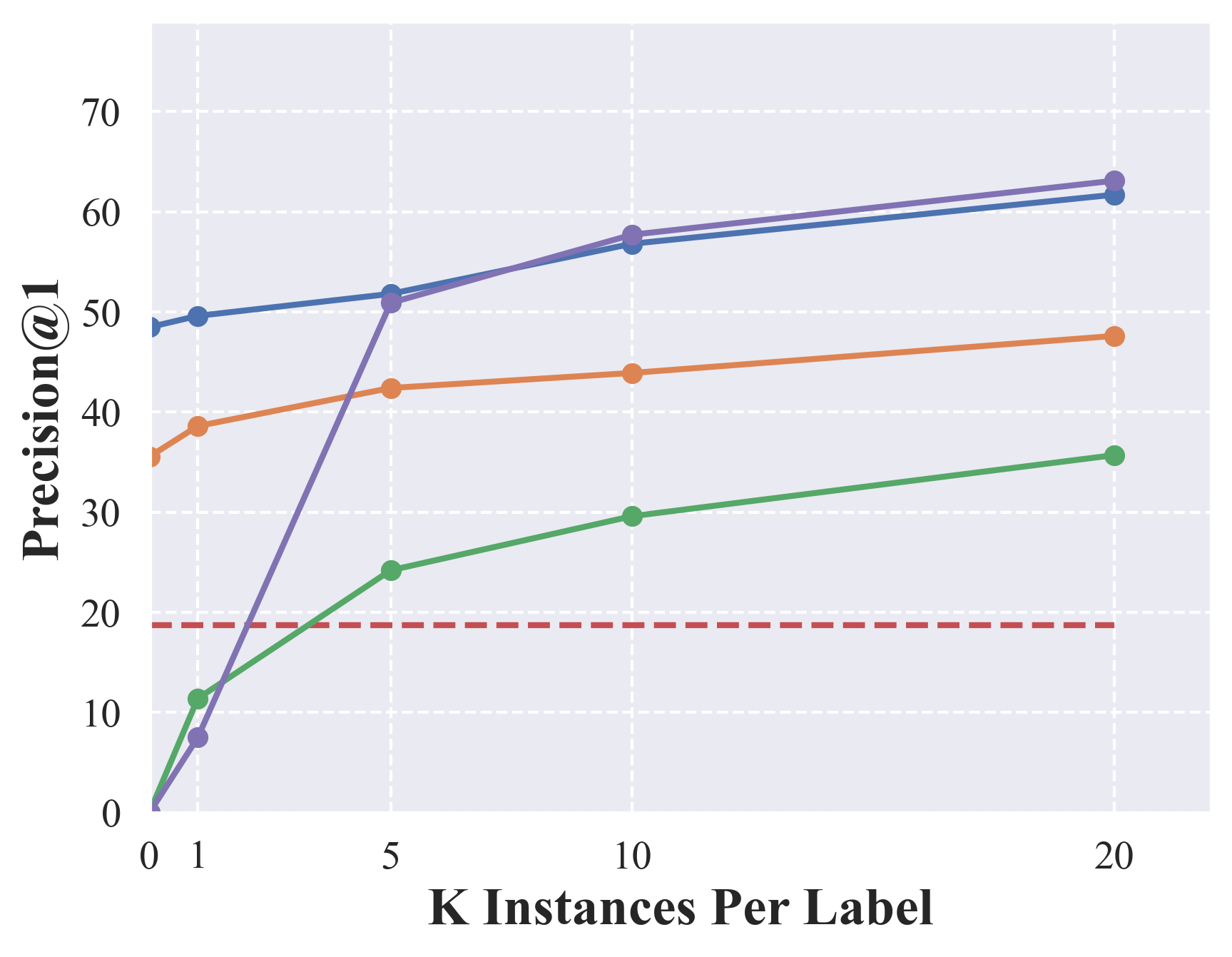}
  \caption{\amazonfull{}}
  \label{fig:fsp1_amazon}
\end{subfigure}
\caption{
Few-Shot \precision{1} for different Values of K on \eurlex{} and \amazon{}. \semsupxc{} starts off significantly higher and for \eurlex{} maintains the gap for larger values of K to the second best model, MACLR~\cite{Xiong2022ExtremeZL}. For \amazon{}, \semsupxc{} maintains similar leads for most baselines, while being at par with \lightxml{}~\cite{Jiang2021LightXMLTW}. 
% \todo{Pranjal: The above is true only for \eurlex{}. Be careful with such statements. Reflect to make sure it is correct.}
}
\label{fig:few_shot}
\end{figure*}

%% file: LaTeX/600_results.tex
\section{Results}
\label{sec:results}

% In this section, we report results of our models and baselines. We show significant improvements in both Zero-Shot and Few-Shot Settings.

% Zero-shot results
\input{LaTeX/Result_files/00_zero_shot.tex}
\input{LaTeX/Ablation_tables/ablation_table_2}

% Few-shot results
\input{LaTeX/Result_files/01_few_shot.tex}

%% file: LaTeX/Result_files/00_zero_shot.tex
% Table 2: Zero-shot Learning comparison of different methods.
% \input{LaTeX/Tables/zsl_table3.tex}

\subsection{Zero-shot extreme classification}
\label{sec:results:zero_shot}

For the zero-shot scenario, we compare \semsupxc{} with baselines which use class descriptions and counterparts which use class names as auxiliary information.
We provide label hierarchy as additional supervision in both cases.
We compare \semsupxc{} with the best variant of each baseline.
Table~\ref{table:results-zsl3} shows that \semsupxc{} significantly outperforms baselines on almost all datasets and metrics, under both zero-shot (\zsxc{}) and generalized zero-shot (\gzsxc{}) settings.
On \zsxc{}, \semsupxc{} outperforms MACLR by over $24$, $12$, and $6$ \precision{1} points on the three datasets respectively, even though MACLR uses XC specific pre-training.
%  (Inverse Cloze Task~\cite{Lee2022DeduplicatingTD} and SimCSE~\cite{Gao2021SimCSESC})
%  , while \semsupxc{} does not.
\semsupxc{} also outperforms GROOV (e.g., over $48$ \precision{1} points on \eurlex{}) which uses a generative T5 model pre-trained on significantly more data than our BERT backbone.
GROOV's unconstrained output space might be one of the reasons for its worse performance.
\semsupxc{}'s semantic understanding of instances and labels stands out against ZestXML which uses sparse non-contextual features with the former consistently scoring twice as higher compared to the latter.
\semsupxc{} consistently outperforms SPLADE~\citep{Formal2021SPLADESL}, a state-of-the-art information retrieval method.
This shows that the straightforward application of IR baselines on XC, even when they are fine-tuned, underperforms.
This is likely because of the multi-label and fine-grained nature of classes coupled with a heavy-tailed distribution.
% settings results in low scores because the XC paradigm is substantially different and requires methodological alterations to address multiple challenges such as multi-label classification, large tail distribution, small label names compared to huge documents in IR.
For most of the datasets and settings, \tfidf{} is competitive with deep baselines. This is because sparse methods often perform better than dense bi-encoders in zero-shot settings~\citep{Thakur2021BEIRAH}, as the latter fail to capture fine-grained information. However, \semsupxc{}'s hybrid lexico-semantic similarity module (\relaxed{}) can perform fine-grained lexical and semantic matching between input instance and description and thus outperforms both sparse and deep methods on all datasets.
% \semsupxc{}'s architecture allows it to capture both dense
 % in descriptions along with capturing deep semantic information, thus resulting in superior ZS and GZS scores. 
\semsupxc{} also outperforms the unsupervised baselines \tfive{} and \senttrans{}, even though they are pre-trained on significantly larger amounts of data than BERT (\tfive{} use $50\times$ compared our base model).

\semsupxc{} also achieves higher recall on \eurlex{} and \amazon{} datasets, beating the best performing baselines by $6$, $18$ \recall{10} points respectively, while being only 3 \recall{10} points less on \wikipedia{}.
% \semsupxc{} beats the best baselines on \eurlex{} and \amazon{} datasets by $6$ and $18$ \recall{10} points, respectively, while being only 3 points less on \wikipedia{}.
% On \wikipedia{}, \semsupxc{} is only 3 \recall{10} points less than \maclr{}.
\semsupxc{} is also the best model for \gzsxc{}.
The margins of improvement are 1-2 \precision{1} points, which are smaller only because \gzsxc{} includes seen labels during evaluation, which are usually large in number.
We refer readers to Appendix~\ref{app:gzs_performance} for more detailed discussion on \gzsxc{} performance.
% includes labels seen during training while evaluation, all methods have higher scores than \zsxc{} and the gaps between different models are smaller, but we see both on precision and recall metrics that~\semsupxc{} again outperforms all the baselines considered by margins of  1-2 precision@1 points.
Table~\ref{tab:results-zsl} in Appendix~\ref{app:zero-shot} contains additional results with more methods and metrics.
Our results show that \semsupxc{} is able to utilize the semantic and lexical information in class descriptions to improve performance significantly, while other baselines hardly improve when using descriptions instead of class names.
\input{LaTeX/Ablation_tables/ablation_table_1}

% \input{LaTeX/Tables/zsl_table2.tex}

% {\color{red}
% \begin{itemize}
%     \item Compare \semsupxc{} with other models
%     \item Specifically compare \semsupxc{} and other models, including trends on different settings.
%     \item \sout{Compare computational speed?}
% \end{itemize}
% }

% %%%%%%%%%% Old %%%%%%%%%%
% On unseen labels, Table~\ref{tab:results-zsl} indicates our framework surpasses all baseline models by significant margins across all datasets. The First three methods for each dataset were not trained on any of the XMC datasets. Therefore, there generalized zero-shot performance is poor. AttentionXML doesn't uses any label features, and treats labels as class ids. Therefore it achieves trivial scores for zero-shot setting. 
% % In GZSL setting also, we achieve SOTA numbers. However, in this setting achieving good precision scores is possible by predicting only the seen labels. Thus, even AttentionXML(which does not use label features) achieve comparable scores.
% %%%%%%%%%% Old %%%%%%%%%%

%% file: LaTeX/Ablation_tables/ablation_table_1.tex
\begin{table*}[!htbp]
\centering
\resizebox{\linewidth}{!}{%
\begin{tabular}{lcccccccccc}
\toprule
\multirow{2}{*}[-6pt]{\large{\bf{Method}}} & \multicolumn{4}{c}{\bf{Components}}  & \multicolumn{3}{c}{\bf{\eurlexfull{}}} & \multicolumn{3}{c}{\bf{\amazonfull{}}}  \\
\cmidrule(lr){2-5} \cmidrule(lr){6-8} \cmidrule(lr){9-11}
& Auxillary & \multirow{2}{*}{Hierarchy} & Exact & Hybrid & \multirow{2}{*}{\precision{1}} & \multirow{2}{*}{\precision{5}} & \multirow{2}{*}{\recall{10}} & \multirow{2}{*}{\precision{1}} & \multirow{2}{*}{\precision{5}} & \multirow{2}{*}{\recall{10}}
\\
& Information &  & Match & Match &  &  &  &  & 
\\
% \midrule 
% \bf{\semsupxc{}}   & \cmarkgreen & \cmarkgreen & \xmarkred & \cmarkgreen  & $44.7$ & $\mathbf{20.9}$ & $\mathbf{57.4}$ & $\bf{48.2}$ & $\bf{27.0}$ & $\bf{72.9}$  \\
\midrule
\rowcolor{gray!20} \multicolumn{11}{l}{Ablating Label Descriptions} \\ \midrule
\bf{\semsupxc{}}   & Descriptions & \cmarkgreen & \cmarkgreen & \cmarkgreen  & $44.7$ & $\mathbf{20.9}$ & $\mathbf{57.4}$ & $\bf{48.2}$ & $\bf{27.0}$ & $\bf{72.9}$  \\
% $\quad$ Remove hierarchy & Descriptions & \cmarkgreen & \xmarkred & \cmarkgreen    & $36.4$ & $16.3$ & $40.2$ & $21.7$ & $13.6$ & $40.3$  \\
$\quad$ \textcolor{darkyellowtable}{Replace} descriptions with names  & \textcolor{darkyellowtable}{Names} & \cmarkgreen & \cmarkgreen & \cmarkgreen & $\mathbf{45.4}$ & $20.6$ & $57.0$ & $43.9$ & $25.4$ & $ 69.7$  \\
$\quad$ \textcolor{darkredtable}{Remove} hierarchy & Descriptions & \xmarkred & \cmarkgreen & \cmarkgreen    & $30.2$ & $14.2$ & $40.2$ & $21.7$ & $13.6$ & $40.3$  \\
\midrule
\rowcolor{gray!20} \multicolumn{11}{l}{Ablating Model Architecture Components} \\ \midrule
\bf{\semsupxc{}}   & Descriptions & \cmarkgreen & \cmarkgreen & \cmarkgreen  & $\mathbf{44.7}$ & $\mathbf{20.9}$ & $\mathbf{57.4}$ & $\bf{48.2}$ & $\bf{27.0}$ & $\bf{72.9}$  \\
$\quad$ \textcolor{darkyellowtable}{Replace} Hybrid with Exact Lexical Matching & Descriptions & \cmarkgreen & \cmarkgreen & \xmarkred & $42.6$ & $19.3$ & $53.7$ & $45.8$ & $25.5$ & $ 69.2$  \\
% $\quad$ Replace \relaxed{} with \coil{} & Web & \cmarkgreen & \cmarkgreen & \xmarkred & $42.6$ & $19.3$ & $53.7$ & $45.8$ & $25.5$ & $ 69.2$  \\
$\quad$ \textcolor{darkredtable}{Remove} all lexical matching & Descriptions & \cmarkgreen & \xmarkred & \xmarkred & $11.9$ & $8.9$ & $29.4$ & $37.3$ & $22.0$ & $60.6$  \\
\bottomrule
\end{tabular}
% Resize
}
\caption{\label{results-ablation1}
Component-wise Model Analysis of \semsupxc{} for \zsxc{} on \eurlex{} and \amazon{}.
Each component contributes to the final performance, with lexical-matching playing an important role.}
\label{tab:results-ablation1}
\end{table*}

%% file: LaTeX/Ablation_tables/ablation_table_2.tex
% Table 2: Extreme Zero-shot Learning (EZ-XMC) comparison of different unsupervised methods.
\begin{table}[t]
\centering
\resizebox{\columnwidth}{!}{%
\begin{tabular}{lcccccc}
\toprule
\multirow{2}{*}{\bf{Method}} & \multicolumn{3}{c}{\bf{\eurlexfull{}}} & \multicolumn{3}{c}{\bf{\amazonfull{}}}  \\
\cmidrule(lr){2-4} \cmidrule(lr){5-7}
 & \bf{\precision{1}} & \bf{\precision{5}} & \bf{\recall{10}} & \bf{\precision{1}} & \bf{\precision{5}} & \bf{\recall{10}} 
 \\
\midrule
% \semsupxc{}                   & $44.7$\tablestd{2.3} & $20.9$\tablestd{0.6} & $57.4$\tablestd{3.4} & $\bf{48.2}$\tablestd{0.5} & $\bf{27.0}$\tablestd{0.5} & $\bf{72.9}$\tablestd{0.5}  \\
% $\quad$ + Augmentation & $\bf{45.5}$\tablestd{1.6} & $\bf{21.6}$\tablestd{1.1} & $\bf{59.0}$\tablestd{1.9} & $47.8$\tablestd{0.2} & $26.8$\tablestd{0.8} & $72.6$\tablestd{0.3}  \\
\semsupxc{}                   & $44.7$ & $20.9$ & $57.4$ & $\bf{48.2}$ & $\bf{27.0}$ & $\bf{72.9}$  \\
$\quad$ + Augmentation & $\bf{45.5}$ & $\bf{21.6}$ & $\bf{59.0}$ & $47.8$ & $26.8$ & $72.6$  \\
\bottomrule
\end{tabular}
% % Resize
}
\caption{
Description augmentation helps boost performance for \zsxc{} on \eurlex{}, but does not help on \amazon{}, which is a significantly larger dataset ($3\times$  labels).
This demonstrates \semsupxc{}'s out-of-the-box performance, since augmentation is unnecessary for larger label spaces.
% \textcolor{red}{Results ready at 8 PM EST}
}
\label{tab:results-ablation_aug}
\end{table}

%% file: LaTeX/Result_files/01_few_shot.tex
\vspace{-5pt}
\subsection{Few-shot extreme classification}
\label{sec:results:few_shot}

% \pranjal{UPDATED \sout{Talk about LightXML}}

% Figures

% TODO: Fix this hack
% \input{LaTeX/Ablation_tables/ablation_table_1}

% % Table
% \input{LaTeX/Tables/few_shot_table.tex}

We now consider the \fsxc{} setup, where new classes added at evaluation time have a small number of labeled instances each ($K$).
We evaluate on four settings -- $K \in \{ 1, 5, 10, 20 \}$ and all baselines other than \zest{}, which cannot be used for \fsxc{} (See Appendix~\ref{app:few-shot}). Further, we omit evaluation on \wikipedia{}, since it has $\approx 10$ training examples per label, which is insufficient to study the effect of increasing values of $K$.
% \footnote{Such Labels can neither be considered as positives nor negatives. \zest{} cannot be directly modified to handle this; therefore, we omit its results. See Appendix~\ref{app:few-shot} for more details.}
% ).
% Since XC is a multi-label problem, some labels must be dropped (for particular documents) to ensure each label corresponds to exactly K input documents.
For the sake of completeness, we also include zero-shot performance (\zsxc{}, $K=0$) and report results in Figure~\ref{fig:few_shot}.
Detailed results for other metrics (showing the same trend as \precision{1}) and implementation details regarding creation of the few-shot splits are in appendix~\ref{app:few-shot}.

Similar to the \zsxc{} case, \semsupxc{} outperforms all baselines for all values of $K$ on \eurlex{}. For \amazon{}, \semsupxc{} outperforms all baselines other than \lightxml{}. \lightxml{} is significantly outperformed for $K=\{0, 1\}$ and matches for $K=\{5, 10, 20\}$.
% For $K = 20$, \semsupxc{} outperforms baselines on \eurlex{}, but is only 1 \precision{1} points lower than \lightxml{} on \amazon{}. which is a fully supervised baseline, and otherwise performs poorly on zero-shot and lower values of $K$.
In comparison to \maclr{} and \groov{}, \semsupxc{} consistently outperforms by large margins (eg., 12 \& 27\precision{1} points for \eurlex{}) across all values of $K$.
% As expected, \semsupxc{}'s performance increases with $K$ because of access to more labeled data, but crucially, it continues to outperform baselines by the same margins.
\semsupxc{}'s zero-shot performance is higher than even the few-shot scores of \maclr{} and \groov{} that have access to $K=20$ labeled samples on~\amazon{}, which further strengthens the model's applicability to the XC paradigm.
Moreover, adding a few labeled examples seems to be more effective in \eurlex{} than \amazon{}, with the performance difference between $K=1$ and $K=20$ being 22 and 12 \precision{1} points respectively.
% Combined with the fact that performance seems to plateau for both datasets, we believe that the larger label space with rich descriptions for~\amazon{} has allowed~\semsupxc{} to learn label semantics better than for~\eurlex{}.
This, along with the fact that performance seems to plateau for both datasets, suggests that \semsupxc{} learns label semantics better for \amazon{} than \eurlex{}, due to its larger label space with rich descriptions.

% \input{LaTeX/Ablation_tables/ablation_table_2}

% In this setup, we finetune the pretrained model from the previous section on few instances of each of unseen labels. Precisely, we randomly sample instances from the training set, such that each unseen label has exactly k training instances to it. For labels, which have more than k input documents, we do not sample it as positive for the extra input documents. Same procedure is applied for other baselines as well.

%%%%%%%%%% Old %%%%%%%%%%
% \begin{figure}
%     \begin{center}
%     \includegraphics[scale=0.6]{LaTeX/few_shot_figures/eurlex_Precision@1.png}
%     \end{center}
%     \caption{Eurlex Few-Shot Precision@1 for different values of K. }
%     \label{fig:fsp1_eurlex}
% \end{figure}

% \begin{figure}
%     \begin{center}
%     \includegraphics[scale=0.6]{LaTeX/few_shot_figures/eurlex_Precision@5.png}
%     \end{center}
%     \caption{Eurlex Few-Shot Precision@5 for different values of K. }
%     \label{fig:fsp5_eurlex}
% \end{figure}
%%%%%%%%%% Old %%%%%%%%%%

%% file: LaTeX/700_ablations.tex
\subsection{Analysis}
\label{sec:results:ablations}

We dissect the performance of~\semsupxc{} by conducting ablation studies on model components and label descriptions and further provide qualitative analysis on \eurlex{} and \amazon{} for the zero-shot extreme classification setting (\zsxc{}) in the following sections.
% \todo{Should we say why we don't use \wikipedia{}?}
% We do not use \wikipedia{} because of compute constraints.

% Analyzing components
\input{LaTeX/Ablation_files/00_analyzing_components.tex}

% Data augmentation
\input{LaTeX/Ablation_files/01_augmentation}

\input{LaTeX/Ablation_files/01_5_label_sources}

% Qualitative analysis
\input{LaTeX/Ablation_files/02_qualitative_analysis.tex}

% Computational efficiency

% %%%%%%%% Old %%%%%%%%
% \subsection{Description Collection, Hierarchy, Processing}
% \begin{itemize}
%     \item GPT vs Web
%     \item Effect of Hierarchy
%     \item Effect of Heuristics for filtering
%     \item Effect of Processing ( In Wiki dataset)
% \end{itemize}

% \subsection{Choosing Number of shortlists/sampling}
% Number and method of shortlisting and sampling can have some effect on scores, however proper ablation for this will be time-consuming.

% \subsection{Unimportant Ablations}
% \subsubsection{Cross Encoder}
% \note{Unlikely to use.}
% \subsubsection{Effect of Output Encoder}
% Can discuss size, freezing of initial layers of output encoder, though might be unecessary
% %%%%%%%% Old %%%%%%%%

%% file: LaTeX/Ablation_files/00_analyzing_components.tex
\paragraph{Ablating components of \semsupxc{}}
\input{LaTeX/Ablation_tables/ablation_table_3}

\input{LaTeX/Ablation_tables/qualitative_table}

% Table

\semsupxc{}'s use of the~\relaxed{} model and semantically rich descriptions enables it to outperform all baselines considered, and we analyze the importance of each component in Table~\ref{tab:results-ablation1}.
As our base model (first row) we consider~\semsupxc{} without ensembling it with~\tfidf{}.
We note that the~\semsupxc{} base model is the best performing variant for both datasets and on all metrics other than \precision{1} for \eurlex{}, for which it is only $0.7$ points lower.
Web scraped class descriptions are important because removing them decreases both precision and recall scores (e.g., \precision{1} is lower by $4$ points on \amazon{}) on all settings considered.
We see bigger improvements with \amazon{}, which is the dataset with larger number of classes (13K), which substantiates the need for semantically rich descriptions when dealing with a large number of fine-grained classes.
Label hierarchy information is similarly crucial, with large performance drops on both datasets in its absence (e.g., $26$ \precision{1} points on \amazon{}), thus showing that access to structured hierarchy information leads to better semantic representations of labels.

On the modeling side, we observe that relaxed and exact lexical matching, which are components of \relaxed{}, are important, with their absence leading to $2$ and $11$ \precision{1} point degradation on \amazon{}.
Even for \eurlex{}, hybrid lexical matching improves performance by $33$ \precision{1} points when compared to a model with no lexical matching.
% over exact lexical-matching is very important.
This highlights that our proposed model \relaxed{}'s hybrid semantic-lexical approach significantly improves performance  on \xc{} datasets.
% both smaller and larger \xc{} datasets.
%  significantly more important for \eurlex{} which is the smaller dataset, we see that both types of matching are important for \amazon{} which tends to have classes which are more related.
% \pranjal{Can we add examples of classes for Amazon which substantiate the above?}

%% file: LaTeX/Ablation_tables/ablation_table_3.tex
% Table 2: Extreme Zero-shot Learning (EZ-XMC) comparison of different unsupervised methods.
\begin{table}[t]
\centering
\resizebox{\columnwidth}{!}{%
\begin{tabular}{lcccccc}
\toprule
\multirow{2}{*}{\bf{Method}} & \multicolumn{3}{c}{\bf{\eurlexfull{}}} & \multicolumn{3}{c}{\bf{\amazonfull{}}}  \\
\cmidrule(lr){2-4} \cmidrule(lr){5-7}
 & \bf{\precision{1}} & \bf{\precision{5}} & \bf{\recall{10}} & \bf{\precision{1}} & \bf{\precision{5}} & \bf{\recall{10}} 
 \\
\midrule
% \semsupxc{}                   & $44.7$\tablestd{2.3} & $20.9$\tablestd{0.6} & $57.4$\tablestd{3.4} & $\bf{48.2}$\tablestd{0.5} & $\bf{27.0}$\tablestd{0.5} & $\bf{72.9}$\tablestd{0.5}  \\
% $\quad$ + Augmentation & $\bf{45.5}$\tablestd{1.6} & $\bf{21.6}$\tablestd{1.1} & $\bf{59.0}$\tablestd{1.9} & $47.8$\tablestd{0.2} & $26.8$\tablestd{0.8} & $72.6$\tablestd{0.3}  \\
WordNet & $42.8$ & $20.7$ & $55.0$ & $47.2$ & $26.2$ & $72.2$   \\
GPT-3 (6.7B) & $42.5$ & $20.5$ & $55.9$ & $47.0$ & $26.8$ & $72.8$ \\
\semsupxc{}  & $\bf{44.7}$ & $\bf{20.9}$ & $\bf{57.4}$ & $\bf{48.2}$ & $\bf{27.0}$ & $\bf{72.9}$  \\

\bottomrule
\end{tabular}
% % Resize
}
\caption{
\semsupxc{} significantly outperforms methods using descriptions from alternative sources like WordNet and GPT-3. Our approach consistently improves performance on \eurlex{} and \amazon{} datasets. 
Additionally, our proposed method generates more diverse descriptions, scales efficiently to large datasets unlike LLM-based approaches, and provides descriptions for non-dictionary proper nouns, unlike WordNet.
% \textcolor{red}{Results ready at 8 PM EST}
}
\label{tab:desc_sources}
\end{table}

%% file: LaTeX/Ablation_tables/qualitative_table.tex
\begin{table*}[!t]
\centering
\resizebox{\linewidth}{!}{%
\begin{tabular}{cll}
\toprule
\multirow{2}{*}{\bf{Input Document}} & \multicolumn{2}{c}{\bf{Top 5 Predictions}} \\
\cmidrule{2-3}
& \multicolumn{1}{c}{\textbf{\semsupxcbig{}}} &  \multicolumn{1}{c}{\textbf{\maclr{}}} \\ 
\midrule
\multirow{5}{0.7\linewidth}{Start-Up: A Technician's Guide. In addition to being an excellent stand-alone self-instructional guide, ISA recommends this book to prepare for the Start-Up Domain of CCST Level I, II, and III examinations.} & test preparation & vocational tests \\
& schools \& teaching & graduate preparation \\
& \textbf{new} & test prep \& study guides \\
& \textbf{used and rental textbooks} &  testing\\
& software & vocational  \\
\midrule
\multirow{5}{0.7\linewidth}{Homecoming (High Risk Books). When Katey Bruscke's bus arrives in her unnamed hometown, she finds the scenery blurred, "as if my hometown were itself surfacing from beneath a black ocean." At \ldots}
& \textbf{literature \& fiction} & friendship \\
& thriller \& suspense & mothers \& children\\
& thrillers &  drugs \\
& genre fiction & coming of age\\
& general & braille \\
\midrule
\multirow{5}{0.7\linewidth}{Rolls RM65 MixMax 6x4 Mixer. The new RM65b HexMix is a single rack space unit featuring 6 channels of audio mixing, each with an XLR Microphone Input and 1/4\" unbalanced Line Input. A unique \ldots}
& \textbf{studio recording equipment} & powered mixers\\
& powered mixers & hand mixers \\
& home audio & mixers \& accessories \\
& \textbf{musical instruments} & mixers \\
& speaker parts \& components & mixer parts \\
\bottomrule

\end{tabular}
}
\caption{
\label{table:qualitative}
Sample predictions from \semsupxc{} (our model) compared to MACLR~\citep{Xiong2022ExtremeZL}. Bold represents correct predictions.
Qualitative analysis shows that~\semsupxc{} can understand the document at a higher level than baselines like~\maclr{}.
The second example poses an especially interesting case where~\semsupxc{} is able to understand that the document is a fiction book, whereas~\maclr{} tries to parse the story itself and predicts all labels incorrectly.
}

\end{table*}

%% file: LaTeX/Ablation_files/01_augmentation.tex
\paragraph{Augmenting Label Descriptions}

% Insert the table
% \input{LaTeX/Ablation_tables/ablation_table_2.tex}

% \pranjal{Make sure you add details about data augmentation and cite the relevant techniques. There's no need to be verbose, implementation details can go in the appendix.}

The previous result showed the importance of class descriptions, and we explore the effect of augmenting them to increase their diversity and quantity (See Table~\ref{tab:results-ablation_aug}).
%  to increase their number, and hence \semsupxc{}'s understanding of the class, and 
% We report the results in Table~\ref{tab:results-ablation_aug}.
We use the easy data augmentation (EDA) method~\citep{eda} for augmentations.
Specifically, we apply random word deletion, random word swapping, random insertion, and synonym replacement each with a probability of 0.5 on each description, and add the augmented descriptions to the original ones.
We notice that augmentation improves performance on~\eurlex{} by $1$, $1$, and $2$ \precision{1}, \precision{5}, and \recall{10} points respectively, suggesting that augmentation can be a viable way to increase the quantity of descriptions.
On~\amazon{}, augmentation has no effect on the performance and rather slightly hurts it (e.g., $0.4$ \precision{1} points).
Given that \amazon{} has $3\times$ the number of labels in~\eurlex{}, we believe this shows~\semsupxc{}'s effectiveness in capturing the label semantics in the presence of a larger number of classes, rendering data augmentation redundant.
However, we believe that data augmentation might be a simple tool to boost performance on smaller datasets with lesser labels or descriptions.

%% file: LaTeX/Ablation_files/01_5_label_sources.tex
\subsection{Alternate Sources of Descriptions}

We further assess the impact of utilizing class descriptions from various sources. In Table~\ref{tab:desc_sources}, we compare the performance of descriptions obtained from 1.) \textbf{Language Model generated:} We generate descriptions using variant of GPT-3 with 6.7B parameters, 2.) \textbf{Knowledge Base}: We use definitions provided in WordNet as descriptions, and 3.) \textbf{\semsupxc{}}: Our proposed method. Our method consistently outperforms the others, with a 2 \precision{1} point improvement on \eurlex{} and a 1 \precision{1} point improvement on \amazon{}. Furthermore, unlike LLM-based approaches, our method can efficiently scale to datasets containing millions of labels, such as \wikipedia{}. \semsupxc{} generates more diverse descriptions compared to those available in WordNet and is also applicable to classes containing proper nouns or non-dictionary words. See Appendix~\ref{app:desc_sources} for more details.

%% file: LaTeX/Ablation_files/02_qualitative_analysis.tex
\paragraph {Qualitative analysis}

% We now perform a qualitative analysis of \semsupxc{}'s predictions and present representative examples in Table~\ref{table:qualitative}, and compare them to \maclr{} which is the next best performing model.
We present a qualitative analysis of the performance \semsupxc{}'s predictions in Table~\ref{table:qualitative} compared to \maclr{}.
Examples are instances where \semsupxc{} outperforms \maclr{}, highlighting the strengths of our method, with correct predictions in bold.
In the first example, 
% despite the brevity, \semsupxc{} is able to figure out that it is not just a book, but a \textit{textbook}.
while MACLR predicts five labels which are all similar, \semsupxc{} is able to predict diverse labels while getting the correct label in five predictions.
In the second example, \semsupxc{} realizes the content of the document is a story and hence predicts \textit{literature \& fiction}, whereas \maclr{} predicts classes based on the content of the story instead.
This shows the nuanced understanding of the label space that~\semsupxc{} has learned.
% The third example portrays the semantic understanding of \semsupxc{}'s label space.
% While \maclr{} tries to predict labels like \textit{powered mixers} because of the presence of the word \textit{mixer}, \semsupxc{} is able to understand the text at a high level and predict labels like \textit{studio recording equipment} even though the document has no explicit mention of the words studio, recording or equipment.
In the third example, \semsupxc{} shows a deep understanding of the label space by predicting "studio recording equipment" even though the document has no explicit mention of the words studio, recording or equipment. For same example, \maclr{} fails as it predicts labels like \textit{powered mixers} because of the presence of the word \textit{mixer}.
These examples show that \semsupxc{}'s understanding of how different fine-grained classes are related and how instances refer to them is better than the baselines considered.
We list more such examples in Appendix~\ref{app:qual_analysis}.
% 1.) In first example, one of the correct labels was aerospace, but given from the short description, model predicts reasonably well. Moreover unlike MACLR which keeps predicting almost similar predictions, our model provides diverse outputs (ex used and rental textbooks) which MACLR fails to capture.
% 2.) MACLR predicts what is inside the book. However our model predicts the category in which the book will fall in fullfiling the requirements of dataset.
% 3.) MACLR is highly dependent on text and keeps outputting very straightforward labels. On the other hand, our model outputs labels such as studio recording equipment without the text containing any of it(neither studio, recording or equipment).

%% file: LaTeX/300_related_work.tex
\section{Related Work}
\label{sec:related}

% %%%% Teaser figure %%%%
% \begin{figure*}
% \centering
% \includegraphics[width=\linewidth]{LaTeX/Teaser_figure/Teaser_Figure.pdf}
% \caption{Our model \semsupxc{} improves the framework of semantic supervision~\cite{Hanjie2022SemanticSE}, by adding (1) large-scale automated class description collection with heuristic filtering, (2) contrastive learning, and (3) a novel lexico-semantic matching model building on COIL~\cite{Gao2021COILRE}. While (1) and (2) significantly improve~\semsup{}'s computational speed (over 99\% on~\wikipedia{}), (3) boosts performance~\symbolsecref{sec:results:zero_shot}
% \ameet{Cite COIL in the Figure as well.}
% \ameet{Can we give a couple examples for heuristics?}
% }
% \label{fig:teaser}
% \end{figure*}
% %%%%%%%%%%%%%%%%%%%%%%%

\paragraph{Extreme classification}
% \subsection{Extreme Classification}
Extreme classification (\xc{})~\cite{Agrawal2013MultilabelLW} studies multi-class and multi-label classification problems over numerous classes (thousands to millions).
Traditionally, studies have used \textit{sparse} bag-of-words features of input documents~\citep{Bhatia2015SparseLE, Chang2019AMD, Lin2014MultilabelCV}, simple one-versus-all binary classifiers~\citep{Babbar2017DiSMECDS, Yen2017PPDsparseAP, Jain2019SliceSL,dahiya2021siamesexml}, and tree-based methods which utilize the label hierarchy~\citep{Parabel, Wydmuch2018ANG, Khandagale2020BonsaiDA}.
% 1) Sparse-features based methods generally use bag-of-words representations for input documents.
% While one v.s. all methods (DiSMEC \citet{Babbar2017DiSMECDS}, PPDSparse \citet{Yen2017PPDsparseAP}, Slice \citet{Jain2019SliceSL}), tree-based methods(Parabel \citet{Parabel}, ExtremeText \citet{Wydmuch2018ANG}, Bonsai \citet{Khandagale2020BonsaiDA}) partition the label space into hierarchical  tree structure, and perform layer-wise classification resulting in sub-linear complexity \pranjal{UPDATED: ...short summary...}, embeddings-based methods(SLEEC \citet{Bhatia2015SparseLE}, Slinmer \citet{Chang2019AMD}, FaIE \cite{Lin2014MultilabelCV}) project high dimensional output space to low dimension, and commonly use nearest neighbour search methods to shortlist labels. \pranjal{UPDATED: ... short summary ...}.
Recently, neural-network (NN) based contextual \textit{dense-features} have improved accuracies.
% due to their ability to generate contextual representations.
Studies have experimented with convolutional neural networks~\citep{Liu2017DeepLF}, Transformers~\citep{xtransformer,Jiang2021LightXMLTW,zhang2021fast}, attention-based networks~\citep{You2019AttentionXMLLT}, and shallow networks~\citep{medini2019extreme,mittal2021decaf,dahiya2021deepxml}.
While the aforementioned works show impressive performance when the labels during training and testing are the same, they do not consider the practical zero-shot classification scenario with unseen test labels.

\paragraph{Zero-shot classification}
Zero-shot classification (ZS)~\citep{larochelle2008zero} aims to predict unseen classes not encountered during training by utilizing auxiliary information like class names or prototypes.
Multiple works have attempted ZS for text~\citep{dauphin2014zero,nam2016all,wang2018joint,pappas2019gile,Hanjie2022SemanticSE}, however, they face performance degradation and are computationally expensive due to \xc{}'s large label space.
% , they face performance degradation and are computationally expensive.
ZestXML~\citep{Gupta2021GeneralizedZE} was the first study to attempt ZS extreme classification by projecting \textit{non-contextual} bag-of-words input features close to corresponding label features using a sparsified linear transformation.
% , but this limits them to using non-contextual text representations.
Subsequent works have used NNs to generate contextual representations~\citep{Xiong2022ExtremeZL,Simig2022OpenVE,zhang2022metadata,rios2018few}, with MACLR~\citep{Xiong2022ExtremeZL} adding an XC specific pre-training step and GROOV~\citep{Simig2022OpenVE} using a sequence-to-sequence model to predict novel labels.
However, these works use only label names to represent classes (e.g., the word ``car''), which lack semantic information.
We use semantically rich descriptions~\cite{Hanjie2022SemanticSE}, which coupled with our modeling innovations~\symbolsecref{sec:methodology:semsupxc} achieves state-of-the-art performance on ZS-XC.

%% file: LaTeX/750_computation.tex
\paragraph{Computational Efficiency}
\label{sec:efficiency}

In order to ensure efficient inference, similar to the training process, \semsupxc{} makes predictions on the top 1000 labels shortlisted using \tfidf{}.
The results presented in Table~\ref{table:computation-stats} demonstrate that \semsupxc{} achieves comparable throughput to deep baselines such as \maclr{} and \groov{}, while significantly outperforming them in terms of overall performance.
While \zest{} is significantly faster, \semsupxc{}'s \precision{1} is $2\times$ higher.
While \semsupxc{} requires more storage, the modest 17 GB space it occupies on modern hard drives is inconsequential, especially considering the dataset's scale, which comprises over a million labels.
The storage requirements primarily stem from \semsupxc{}'s \relaxed{} module, which necessitates contextualized representations for each token in the description.
These findings demonstrate that \semsupxc{} strikes the optimal balance between throughput and performance while maintaining practical storage requirements.
% In terms of storage we utilize almost 4 times storage as compared to \maclr{}, as we need to store contextualized token embeddings of each label.
% However the overall storage overhead($\approx$ 17.9GB) is small in comparison to significant improvement in performance and comparable speed.
We provide a more detailed analysis of our method in Appendix~\ref{app:computational_efficiency}

%% file: LaTeX/800_conclusion.tex
\section{Conclusion}
\label{sec:conclusion}

We tackle the task of zero-shot extreme classification (XC) which involves very large label spaces, by using 1) \relaxed{}, which incorporates both semantic similarity at the sentence level and relaxed lexical similarity at the token level, 2) contrastive learning to make training efficient, and 3) semantically rich class descriptions to gain a better understanding of the label space.
% using the framework of semantic supervision~\cite{Hanjie2022SemanticSE} that uses class descriptions instead of label IDs. Our method \semsupxc{} innovates using a combination of contrastive learning, hybrid lexico-semantic matching and \semsupweb{} for automated description collection to train effectively for XC.
We achieve state-of-the-art results on three standard XC benchmarks and significantly outperform prior work. 
Our various ablation studies and qualitative analyses demonstrate the relative importance of our modeling choices.
% , while also providing several ablation studies and qualitative analyses demonstrate the relative importance of our various modeling choices. 
Future work can further improve description quality, and given the strong performance of \relaxed{}, can experiment with better architectures to further push the boundaries of this practical task.

%% file: LaTeX/850_limitations.tex
\section{Limitations}

While our method, \semsupxc{}, exhibits promising results, it is important to acknowledge its inherent limitations. Firstly, our approach relies on scraping descriptions from search engines. Although we have implemented post-processing techniques to filter out toxic content and retain only the most relevant search hits, it is possible that biases and harmful elements present in the original data may persist in the scraped descriptions. Furthermore, despite evaluating our method across diverse domains such as legal, shopping, and Wikipedia, it is essential to note that our approach may encounter challenges when applied to datasets where scraping descriptions is not a straightforward task or necessitates specialized technical knowledge that may not be readily available on the web.

%% file: LaTeX/900_appendix_a.tex
\section*{Appendices}

\section{Training Details}
\label{app:training}

% Fix the section/subsections once we start writing it %

\subsection{Hyperparameter Tuning}
\label{app:hyperparameter}
We tune the learning rate, batch\_size using grid search. For the \eurlex{} dataset, we use the standard validation split for choosing the best parameters. We set the input and output encoder's learning rate at $5e^{-5}$ and $1e^{-4}$, respectively. We use the same learning rate for the other two datasets. We use batch\_size of 16 on \eurlex{} and 32 on \amazon{} and \wikipedia{}. For Eurlex, we train our zero-shot model for fixed 2 epochs and the generalized zero-shot model for 10 epochs. For the other 2 datasets, we train for a fixed 1 epoch. For baselines, we use the default settings as used in respective papers. For all daasets, we use same hyperparameters, and for baselines we use comparable settings to \semsupxc{} for fair comparison.

% \subsection{Label Description Collection}

\textbf{Training}

All of our models are trained end-to-end. We use the pretrained BERT model \cite{devlinbert} for encoding input documents, and Bert-Small model \cite{Turc2019WellReadSL} for encoding output descriptions.
For efficiency in training, we freeze the first two layers of the output encoder. We use contrastive learning to train our models and sample hard negatives based on TF-IDF features. 
All implementation was done in PyTorch and Huggingface transformer and experiments were run NVIDIA RTX2080 and NVIDIA RTX3090 gpus.

\subsection{Baselines}
\label{app:baselines}
We use the code provided by \zest{}, \maclr{} and \groov{} for running the supervised baselines. 
We employ the exact implementation of \tfidf{} as used in \zest{}. 
We evaluate \tfive{} as an NLI task~\citep{Xue2021mT5AM}. 
We separately pass the names of each of the top 100 labels predicted by \tfidf{}, and rank labels based on the likelihood of entailment.
We evaluate \senttrans{}  by comparing the similarity between the emeddings of input document and the names of the top 100 labels predicted by \tfidf{}.  
Splade is a sparse neural retreival model that learns label/document sparse expansion via a Bert masked language modelling head.
We use the code provided by authors for running the baselines.
We experiment with various variations and pretrained models, and find splade\_max\_CoCodenser pretrained model with low sparsity($\lambda_d = 1e-6$ \& $\lambda_q = 1e-6$) to be performing the best.
\section{Label descriptions from the web}
\label{app:descriptions}

\subsection{Automatically scraping label descriptions from the web}

We mine label descriptions from web in an automated end-to-end pipeline. We make query of the form `what is <class\_name>'(or component name in case of Wikipedia) on duckduckgo search engine. Region is set to United States(English), and advertisements are turned off, with safe search set to moderate. We set time range from 1990 uptil June 2019. On average top 50 descriptions are scraped for each query. To further improve the scraped descriptions, we apply a series of heuristics:
\begin{itemize}
    \item We remove any incomplete sentences. Incomplete sentences do not end in a period or do not have more than one noun, verb or auxiliary verb in them. \\ Eg: \underline{Label} = {\bf{Adhesives}} ; \underline{Removed Sentence} = \textit{What is the best glue or gel for applying}
    \item Statements with lot of punctuation such as semi-colon were found to be non-informative. Descriptions with more than 10 non-period punctuations were removed. \\
    Eg: \underline{Label} = {\bf{Plant Cages \& Supports}} ; \underline{Removed Description} = \textit{Plant Cages \& Supports. My Account; Register; Login; Wish List (0) Shopping Cart; Checkout \$ USD \$ AUD THB; R\$ BRL \$ CAD \$ CLP \$ \ldots}
    \item We used regex search to identify urls and currencies in the text. Most of such descriptions were spam and were removed. \\ Eg: \underline{Label} = {\bf{Accordion Accessories}} ; \underline{Removed Description} = \textit{Buy Accordion Accessories Online, with Buy Now \& Pay Later and Rental Options. Free Shipping on most orders over \$250. Start Playing Accordion Accessories Today!}
    \item Descriptions with small sentences(<5 words) were removed. \\ Eg: \underline{Label} = {\bf{Boats}} ; \underline{Removed Description} = \textit{Boats for Sale. Buy A Boat; Sell A Boat; Boat Buyers Guide; Boat Insurance; Boat Financing ...}
    \item Descriptions with more than 2 interrogative sentences were filtered out. \\
    Eg: \underline{Label} = {\bf{Shower Curtains}} ; Removed Description = \textit{So you're interested...why? you're starting a company that makes shower curtains? or are you just fooling around? Wiki User  2010-04} 
    \item We mined top frequent n-grams from a sample of scraped descriptions, and based on it identified n-grams which were commonly used in advertisements. Examples include: \textit{`find great deals', `shipped by'}.
    \\ \underline{Label} = {\bf{Boat compasses}} ; Removed Description = \textit{{\bf{Shop and read reviews}} about Compasses at West Marine. {\bf{Get free shipping}} on all orders to any West Marine Store {\bf{near you today}}}.
    \item We further remove obscene words from the datasets using an open-source library \citep{profanity}.
    \item We also run a spam detection model \citep{spamdet} on the descriptions and remove those with a confidence threshold above 0.9. \\
    Eg: \underline{Label} = {\bf{Phones}} ; \underline{Removed Description} = \textit{Check out the Phones page at <xyz\_company> — the world's leading music technology and instrument retailer!} 
    \item Additionally, most of the sentences in first person, were found to be advertisements, and undetected by previous model. We remove descriptions with more than 3 first person words (such as I, me, mine) were removed. \\
    Eg: \underline{Label} = {\bf{Alarm Clocks}} ; \underline{Removed Description} = \textit{We selected the best alarm clocks by taking the necessary, well, time. We tested products with our families, waded our way through expert and real-world user opinions, and determined what models lived up to manufacturers' claims. \ldots}
    % Sweetwater
\end{itemize}

\subsection{Post-Processing}
\label{app:desc_hier}
We further add hierarchy information in a natural language format to the label descriptions for \amazon{} and \eurlex{} datasets. Precisely, we follow the format of `{key} is {value}.' with each key, value pair represented in new line.
Here key belongs to the set \{ `Description', `Label', `Alternate Label Names', `Parents', `Children' \}, and the value corresponds to comma separated list of corresponding information from the hierarchy or scraped web description.
For example, consider the label `video surveillance' from \eurlex{} dataset. 
We pass the text: \textit{\\`Label is video surveillance.\\Description is <web\_scraped\_description>.\\Parents are video communications.\\Alternate Label Names are camera surveillance, security camera surveillance.'}\\ to the output encoder. \\
For \wikipedia{}, label hierarchy is not present, so we only pass the description along with the name of label.

\subsection{Wikipedia Descriptions}
\label{app:desc_wiki}
When labels are fine-grained, as in the Wikipedia dataset, making queries for the full label name is not possible.
For example, consider the label `Fencers at the 1984 Summer Olympics' from Wikipedia categories; querying for it would link to the same category on Wikipedia itself. 
Instead, we break the label names into separate constituents using a dependency parser. 
Then for each constituent(`Fencers' and `Summer Olympics'), we scrape descriptions.
No descriptions are scraped for constituents labelled by Named-Entity Recognition(`1984'), and their NER tag is directly used.
Finally, all the scraped descriptions are concatenated in a proper format and passed to the output encoder. \\

\subsection{De-Duplication}
To ensure no overlap between our descriptions and input documents, we used SuffixArray-based exact match algorithm ~\citep{Lee2022DeduplicatingTD} with a minimum threshold of 60 characters and removed the matched descriptions.

\section{\relaxed{}}
\label{app:hybridImplementation}
We propose \relaxed{} to exploit token similarity. We create clusters of tokens based on: 1) the BERT \textit{token-embedding similarity}~\citep{Rajaee2021ACAisotropic} is higher than a threshold or 2) if tokens share the same lemma. Specifically, first tokens with BERT embedding cosine similarity greater than 0.6 are put into same cluster. In the second stage if two different tokens share the same lemma, but are in different clusters, their clusters are merged. In model, a mask is created of size (Q * LQ * D * LD), where Q is the number of label descriptions, LQ is the max length of all label descriptions, D is the number of documents, and LD is the length of label descriptions. Here a entry of 1 means that corresponding token in label description and input share the same cluster, else it is set to 0.

\section{Contrastive Learning}
\label{app:contrastive}
During training, for both \eurlex{} and \amazon{}, we sample $1000 - \vert Y_i^{+} \vert$ hard negative labels for each input document. 
For \wikipedia{}, we precompute the top 1000 labels for each input based on \tfidf{} scores.
We then randomly sample $1000 - \vert Y_i^{+} \vert$ negative labels for each document.
At inference time, we evaluate our models on all labels for both \eurlex{} and \amazon{}.
However, even evaluation on millions of labels in \wikipedia{} is not computationally tractable.
Therefore, we evaluate only on top 1000 labels predicted by \tfidf{} for each input.

\section{Full results for zero-shot classification}
\label{app:zero-shot}
\subsection{Split Creation}
\label{app:splitcreation}
For \eurlex{}, and \amazon{}, we follow the same procedure as detailed in \groov{}~\citep{Simig2022OpenVE}. We randomly sample k labels from all the labels present in train set, and consider the remaining labels as unseen. For \eurlex{} we have roughly 25\%(1057 labels) and for \amazon{} roughly 50\%(6500 labels) as unseen. For \wikipedia{}, we use the standard splits as proposed in \zest{}~\citep{Gupta2021GeneralizedZE}.

\subsection{Results}

Table~\ref{tab:results-zsl} contains complete results for \zsxc  across the three datasets, including additional baselines and metrics.

\input{LaTeX/Tables/zsl_table.tex}

\input{LaTeX/Tables/computation_table.tex}

\section{Full results for few-shot classification}
\label{app:few-shot}

\subsection{Split Creation}
We iteratively select k instances of each label in train documents. If a label has more than k documents associated with it, we drop the label from training(such labels are not sampled as either positives or negatives) for the extra documents. We refer to these labels as neutral labels for convenience. Because of such labels, loss functions of dense methods need to be modified accordingly. For \zest{}, this is not possible because it directly learns a transformation over the whole dataset, and individual labels for particular instances cannot be masked as neutral.

\subsection{Models}
We use \maclr{}, \groov{}, \lightxml{} as baselines. 
We initialize the weights from the corresponding pre-trained models in the GZSL setting.
We use the default hyperparameters for baselines and \semsup{} models. 
As discussed in the previous section, neutral labels are not provided at train time for \maclr{} and \groov{} baselines.
However, since \lightxml{} uses a final fully-connected classification layer, we cannot selectively remove them for a particular input. Therefore, we mask the loss for labels which are neutral to the documents. We additionally include scores for \tfidf{}, but since it is a fully unsupervised method, only zero-shot numbers are included. 

\subsection{Results}

The full results for few-shot classification are present in Table~\ref{table:few_shot}.

% Table
\input{LaTeX/Tables/few_shot_table.tex}

\input{LaTeX/Appendix_files/004_computatioal_efficiency}
\input{LaTeX/Appendix_files/006_oracle_table}

\input{LaTeX/Tables/007_d_unseen_table}

\input{LaTeX/Appendix_files/05_qualitative_analysis}

\section{Analysis}

\input{LaTeX/Appendix_files/009_label_sources}

\subsection{Performance on the \gzsxc{} Split}
\label{app:gzs_performance}

To gain further insights into the higher performance on the \gzsxc{} split compared to the \zsxc{} split, we conducted additional evaluations. In Table~\ref{tab:oracle_table}, we compare a hypothetical method denoted as $Oracle_{Seen}$, which achieves perfect accuracy in predicting seen classes while completely avoiding predictions for unseen classes. Although such a high level of prediction capability is impractical in real-world scenarios, the significant advantage demonstrated by $Oracle_{Seen}$ highlights that competitive scores on the \gzsxc{} split can be attained even by disregarding unseen classes. This is not favourable, therefore it is crucial to consider and compare both \gzsxc{} and \zsxc{} scores when evaluating and comparing different methods. It is worth noting that this also explains the \textit{relatively} smaller improvement achieved by \semsupxc{} over other methods in comparison to the \zsxc{} split. Nonetheless, \semsupxc{} consistently outperforms all other baselines across various metrics and settings, showcasing its superior understanding of both seen and unseen labels.

Furthermore, when assessing the Recall@10 metric, it is important to observe that \semsupxc{} is the only method that achieves statistically significant margins over $Oracle_{Seen}$ on the \eurlex{} and \amazon{} datasets, with improvements of 2.9 and 16.7 points, respectively. This outcome can be attributed to the fact that achieving a higher Recall@10 score necessitates a broader coverage of correct labels in the predictions, which is limited to seen labels only. Therefore, only a method that can effectively classify both seen and unseen labels simultaneously, such as \semsupxc{}, can achieve a higher Recall@10 value.

To provide further evidence that methods like \groov{} achieve high performance on the \gzsxc{} split solely by predicting seen labels without truly understanding unseen labels, we introduce a new metric, denoted as $P(D_{unseen}, Y)$. In this metric, Y represents the gold labels as before, and $D_{unseen}$ indicates that only the model's predictions on unseen labels are taken into account. We evaluate different methods using this metric specifically on the \gzsxc{} split. Intuitively, higher scores on this metric indicate the extent to which unseen labels contribute to achieving a high score on the \gzsxc{} split.

As we can observe from Table~\ref{tab:d_unseen_table}, both \groov{} and \zest{} perform poorly on this metric, indicating their limited understanding of unseen labels. Conversely, \maclr{} demonstrates decent performance, while \semsupxc{} emerges as the best performing method by significant margins. This finding suggests that \semsupxc{}'s higher performance on the \gzsxc{} split is a result of effectively considering and classifying both seen and unseen labels, rather than solely relying on seen labels.

%% file: LaTeX/Tables/zsl_table.tex
% Table 2: Extreme Zero-shot Learning (EZ-XMC) comparison of different unsupervised methods.
\begin{table*}[t]
\centering
\resizebox{\linewidth}{!}{%
\begin{tabular}{l|ccc|cc|ccc}
\hline
\multirow{2}{4em}{Method} & \multicolumn{3}{c}{ Precision - ZSL } & \multicolumn{2}{|c}{ Recall - ZSL / GZSL } & \multicolumn{3}{|c}{ Precision - GZSL } \\\cmidrule(lr){2-4}\cmidrule(lr){5-6}\cmidrule(lr){7-9}
 & $@1$ & $@3$ & $@5$ & $R@10$ & $R@10$ & $@1$ & $@3$ & $@5$   \\
\hline

\multicolumn{9}{c}{ Eurlex-4.3K } \\
\hline 
\finalnum{TF-IDF} & $44.0$\tablestd{0.0} & $26.9$\tablestd{0.0} & $19.6$\tablestd{0.0} & $55.8$\tablestd{0.0} & $41.2$\tablestd{0.0} & $53.4$\tablestd{0.0} & $35.2$\tablestd{0.0} & $28.0$\tablestd{0.0} \\
\finalnum{T5} & $7.2$\tablestd{0.0} & $7.1$\tablestd{0.0} & $7.0$\tablestd{0.0} & $29.2$\tablestd{0.0} & $23.0$\tablestd{0.0} & $10.4$\tablestd{0.0} & $11.0$\tablestd{0.0} & $11.2$\tablestd{0.0} \\
\finalnum{Sentence Transformer} & $15.9$\tablestd{0.0} & $10.8$\tablestd{0.0} & $9.1$\tablestd{0.0} & $31.1$\tablestd{0.0} & $25.5$\tablestd{0.0} & $18.8$\tablestd{0.0} & $15.7$\tablestd{0.0} & $11.9$\tablestd{0.0} \\
\hline
% \finalnum{LightXML} & $-$\tablestd{0.0} & $-$\tablestd{0.0} & $-$\tablestd{0.0} & $-$\tablestd{0.0} & $55.2$\tablestd{0.0} & $87.4$\tablestd{0.0} & $67.3$\tablestd{0.0} & $50.5$\tablestd{0.0} \\
\finalnum{ZestXML} &  $9.6$\tablestd{0.0} & $7.3$\tablestd{0.0} & $6.5$\tablestd{0.0} & $25.7$\tablestd{0.0} & $54.8$\tablestd{0.0} & $84.8$\tablestd{0.0} & $64.8$\tablestd{0.0} & $48.9$\tablestd{0.0} \\
\finalnum{ZestXML + TF-IDF} & $24.7$\tablestd{0.0} & $17.7$\tablestd{0.0} & $14.4$\tablestd{0.0} & $46.4$\tablestd{0.0} & $54.2$\tablestd{0.0} & $84.9$\tablestd{0.0} & $65.7$\tablestd{0.0} & $50.3$\tablestd{0.0} \\
\finalnum{MACLR} & $24.9$\tablestd{0.6} & $16.6$\tablestd{0.2} & $13.4$\tablestd{0.2} & $42.1$\tablestd{0.5} & $55.2$\tablestd{1.3} & $60.7$\tablestd{1.3} & $49.1$\tablestd{1.9} & $41.1$\tablestd{1.3} 
\\
\finalnum{GROOV} & $1.2$ \tablestd{0.1} & $2.6$ \tablestd{0.4} & $2.6$ \tablestd{0.3}  & $7.0$ \tablestd{0.9} & $49.4$ \tablestd{0.1} & $84.1$ \tablestd{0.1} & $61.5$ \tablestd{0.2} & $45.3$ \tablestd{0.1} 
\\

\hline
\finalnum{\semsupxc{}-Hier} & $45.4$\tablestd{0.2} & $28.1$\tablestd{0.1} & $20.6$\tablestd{0.2} & $57.0$\tablestd{0.2} & $65.6$\tablestd{1.0} & $86.4$\tablestd{0.2} & $\mathbf{69.2}$\tablestd{0.2} & $\mathbf{54.2}$\tablestd{0.4} \\

\finalnum{\semsupxc{}} & $44.7$\tablestd{2.3} & $27.9$\tablestd{1.1} & $20.9$\tablestd{0.6} & $57.4$\tablestd{3.4} & $\mathbf{65.2}$\tablestd{0.3} & $\mathbf{87.1}$\tablestd{0.1} & $68.5$\tablestd{0.1} & $53.7$\tablestd{0.1} \\

\finalnum{\semsupxc{} + TF-IDF} & $\mathbf{49.3}$\tablestd{0.9} & $\mathbf{31.2}$\tablestd{0.8} & $\mathbf{23.1}$\tablestd{0.3} & $\mathbf{62.4}$\tablestd{0.8} & $62.9$\tablestd{1.1} & $87.0$\tablestd{0.1}  & $67.6$\tablestd{0.1} & $51.6$\tablestd{0.6} \\
\hline \hline

\multicolumn{9}{c}{ Amazon-13K } \\
\hline 
\finalnum{TF-IDF} & $18.7$\tablestd{0.0} & $11.5$\tablestd{0.0} & $8.5$\tablestd{0.0} & $21.0$\tablestd{0.0} & $14.7$\tablestd{0.0} & $21.5$\tablestd{0.0} & $14.4$\tablestd{0.0} & $11.1$\tablestd{0.0} \\
\finalnum{T5} & $2.5$\tablestd{0.0} & $2.8$\tablestd{0.0} & $3$\tablestd{0.0} & $10.5$\tablestd{0.0} & $10.2$\tablestd{0.0} & $3.2$\tablestd{0.0} & $4.2$\tablestd{0.0} & $4.9$\tablestd{0.0} \\
\finalnum{Sentence Transformer} & $15.2$\tablestd{0.0} & $10.5$\tablestd{0.0} & $8.3$\tablestd{0.0} & $22.2$\tablestd{0.0} & $16.0$\tablestd{0.0} & $18.4$\tablestd{0.0} & $13.4$\tablestd{0.0} & $11.0$\tablestd{0.0} \\
\hline
% LightXML & $-$\tablestd{0.0} & $-$\tablestd{0.0} & $-$\tablestd{0.0} & $-$\tablestd{0.0} & $-$\tablestd{0.0} & $X$\tablestd{0.0} & $X$\tablestd{0.0} & $X$\tablestd{0.0} \\
\finalnum{ZestXML} &  $12.7$\tablestd{0.0} & $8.9$\tablestd{0.0} & $7.1$\tablestd{0.0} & $21.2$\tablestd{0.0} & $52.5$\tablestd{0.0} & $87.9$\tablestd{0.0} & $58.6$\tablestd{0.0} & $41.5$\tablestd{0.0} \\
\finalnum{ZestXML + TF-IDF} & $15.6$\tablestd{0.0} & $11.1$\tablestd{0.0} & $8.8$\tablestd{0.0} & $24.4$\tablestd{0.0} & $54.2$\tablestd{0.0} & $87.6$\tablestd{0.0} & $59.0$\tablestd{0.0} & $42.3$\tablestd{0.0} \\
\finalnum{MACLR} & $36.0$\tablestd{0.6} & $23.5$\tablestd{0.4} & $18.0$\tablestd{0.4} & $54.4$\tablestd{0.8} & $46.9$\tablestd{0.4} & $46.0$\tablestd{0.3} & $33.7$\tablestd{0.3} & $27.2$\tablestd{0.3} \\
\finalnum{GROOV} & $0.0$\tablestd{0.0} & $0.3$\tablestd{0.0} & $0.5$\tablestd{0.0} & $2.4$ \tablestd{0.2} & $47.9$\tablestd{0.3} & $87.4$\tablestd{0.5} & $55.8$\tablestd{0.8} & $38.8$\tablestd{0.5} \\

\hline
\finalnum{\semsupxc{}-Hier} & $43.9$\tablestd{0.4} & $31.5$\tablestd{0.7} & $25.4$\tablestd{0.3} & $69.7$\tablestd{0.6} & $71.5$\tablestd{0.3} & $88.4$\tablestd{0.3} & $65.0$\tablestd{0.5} & $50.2$\tablestd{0.3} \\

\finalnum{\semsupxc{} + TF-IDF} & $\mathbf{48.2}$\tablestd{0.5} & $\mathbf{33.9}$\tablestd{0.2} & $\mathbf{27.0}$\tablestd{0.5} & $\mathbf{72.9}$\tablestd{0.5} & $\mathbf{71.6}$\tablestd{0.3} & $\mathbf{88.6}$\tablestd{0.1} & $\mathbf{65.3}$\tablestd{0.3} & $\mathbf{51.2}$\tablestd{0.1} \\

% \semsupxc{} + TF-IDF & $X$ & $X$ & $X$ & $X$ & $71.7$ & $88.5$ & $65.5$ & $51.2$ \\
\hline \hline

\multicolumn{9}{c}{ Wikipedia-1M } \\
\hline 
\finalnum{TF-IDF} & $14.5$\tablestd{0.0} & $7.7$\tablestd{0.0} & $5.5$\tablestd{0.0} & $18.3$\tablestd{0.0} & $14.7$\tablestd{0.0} & $14.4$\tablestd{0.0} & $8.5$\tablestd{0.0} & $6.5$\tablestd{0.0} \\
\finalnum{T5} & $8.2$\tablestd{0.0} & $7.6$\tablestd{0.0} & $6.7$\tablestd{0.0} & $23.6$\tablestd{0.0} & $15.1$\tablestd{0.0} & $4.2$\tablestd{0.0} & $4.5$\tablestd{0.0} & $4.4$\tablestd{0.0} \\
\finalnum{Sentence Transformer} & $19.6$\tablestd{0.0} & $11.1$\tablestd{0.0} & $7.9$\tablestd{0.0} & $22.5$\tablestd{0.0} & $16.6$\tablestd{0.0} & $14.2$\tablestd{0.0} & $9.1$\tablestd{0.0} & $7.0$\tablestd{0.0} \\
\hline
% LightXML & $-$\tablestd{0.0} & $-$\tablestd{0.0} & $-$\tablestd{0.0} & $-$\tablestd{0.0} & $X$\tablestd{0.0} & $X$\tablestd{0.0} & $X$\tablestd{0.0} & $X$\tablestd{0.0} \\
\finalnum{ZestXML} &  $12.9$\tablestd{0.0} & $8.0$\tablestd{0.0} & $6.0$\tablestd{0.0} & $20.0$\tablestd{0.0} & $25.7$\tablestd{0.0} & $26.7$\tablestd{0.0} & $18.8$\tablestd{0.0} & $14.6$\tablestd{0.0} \\
\finalnum{ZestXML + TF-IDF} & $15.8$\tablestd{0.0} & $8.9$\tablestd{0.0} & $6.4$\tablestd{0.0} & $20.8$\tablestd{0.0} & $26.3$\tablestd{0.0} & $30.6$\tablestd{0.0} & $22.2$\tablestd{0.0} & $17.2$\tablestd{0.0} \\
\finalnum{MACLR} & $29.8$\tablestd{0.8} & $17.8$\tablestd{0.6} & $13.2$\tablestd{0.4} & $\mathbf{41.7}$\tablestd{1.3} & $32.7$\tablestd{0.6} & $28.0$\tablestd{0.3} & $18.3$\tablestd{0.3} & $14.4$\tablestd{0.4} \\
\finalnum{GROOV} & $6.0$\tablestd{0.1}& $5.8$\tablestd{0.3} & $5.0$\tablestd{0.4} & $15.4$\tablestd{0.2} & $29.0$\tablestd{0.2} & $31.4$\tablestd{0.1} & $\mathbf{24.9}$\tablestd{0.1} & $\mathbf{19.1}$\tablestd{0.0} \\

\hline
% \semsupxc{}-Names & $X$ & $X$ & $X$ & $X$ & $X$ & $X$ & $X$ & $X$ \\

\semsupxc{} & $34.6$\tablestd{0.2} & $18.9$\tablestd{0.2} & $13.1$\tablestd{0.3} & $37.9$\tablestd{0.4} & $33.0$\tablestd{0.3} & $29.8$\tablestd{0.1} & $22.0$\tablestd{0.2} & $17.0$\tablestd{0.4} \\

\semsupxc{} + TF-IDF & $\mathbf{36.5}$\tablestd{0.3} & $\mathbf{19.5}$\tablestd{0.2} & $\mathbf{13.4}$\tablestd{0.5} & $38.5$\tablestd{0.3} & $\mathbf{34.1}$\tablestd{0.3} & $\mathbf{33.7}$\tablestd{0.4} & $23.4$\tablestd{0.3} & $17.7$\tablestd{0.2} \\
\hline \hline

\end{tabular}
% Bracket close for resizing
}
\caption{\label{results-zsl}
Comparison of \semsupxcbig{} with other supervised and unsupervised baselines. Our method consistently outperfoms all methods across all datasets. 
}
\label{tab:results-zsl}
\end{table*}

%% file: LaTeX/Tables/computation_table.tex
% Table 2: Extreme Zero-shot Learning (EZ-XMC) comparison of different unsupervised methods.
\begin{table}[t]
% \ifx\subformat\iclrformat
\centering
\resizebox{\dimexpr\columnwidth }{!}{%
\begin{tabular}{lccccc}
\toprule
\multirow{2}{*}{\bf{Model}} & \multirow{2}{*}{\textbf{Device}} & \textbf{Throughput}  & \textbf{Storage}  & \textbf{P@1}  \\
 &  & \textbf{(Inputs/s)} & \textbf{(GB)} & \textbf{(ZSL)} \\

\midrule
\semsupxc{} & 1 GPU & $46.2$ &  $17.9$ & $\mathbf{36.5}$ \\
\maclr{} & 1 GPU & $77.8$ & $4.6$ & $29.8$ \\
\groov{} & 1 GPU & $8.9$ & $\mathbf{0.4}$ & $6.0$ \\
\zest{} & 16 CPUs & $\mathbf{2371}$ & $1.8$ & $15.8$ \\
\hline
\end{tabular}}
% \else
% \centering
% \resizebox{\dimexpr\columnwidth - \ifx\subformat\iclrformat \columnwidth / 4 \else 0 \fi}{!}{%
% \begin{tabular}{lccccc}
% \toprule
% \bf{Model} & Device & Throughput (Inputs/s) & Storage (GB) & P@1 (ZSL) \\ 
% \midrule
% \semsupxc{} & 1 GPU & $46.2$ &  $17.9$ & $36.5$ \\
% \maclr{} & 1 GPU & $129.5$ & $4.6$ & $29.8$ \\
% \groov{} & 1 GPU & $8.9$ & $0.4$ & $6.0$ \\
% \zest{} & 16 CPUs & $2371$ & $1.8$ & $15.8$ \\
% \hline
% \end{tabular}}
% \fi

\caption{
Computational Efficiency of \semsupxc{} and baselines on Wikipedia dataset. We have comparable throughput to dense baselines while requiring higher storage but with substantial performance gains.
}
\label{table:computation-stats}
\end{table}

%% file: LaTeX/Tables/few_shot_table.tex
% Table 2: Extreme Zero-shot Learning (EZ-XMC) comparison of different unsupervised methods.
\begin{table*}[t]
\centering
% \resizebox{\columnwidth}{!}{%
\begin{tabular}{lcccccc}
\toprule
\multirow{2}{*}{\bf{Method}} & \multicolumn{3}{c}{\bf{\eurlexfull{}}} & \multicolumn{3}{c}{\bf{\amazonfull{}}}  \\
\cmidrule(lr){2-4} \cmidrule(lr){5-7}
    & \bf{\precision{1}} & \bf{\precision{5}} & \bf{\recall{10}} & \bf{\precision{1}} & \bf{\precision{5}} & \bf{\recall{10}} 
    \\
\midrule
\rowcolor{gray!20} \multicolumn{7}{c}{\bf{1-shot}} \\ \midrule
\semsupxc{} & $\mathbf{50.8}$\tablestd{0.9} & 	$\mathbf{21.4}$\tablestd{0.7} & 	$\mathbf{57.9}$\tablestd{2.9} & 	$\mathbf{49.6}$\tablestd{0.4} & 	$\mathbf{25.2}$\tablestd{0.2} & 	$\mathbf{66.3}$\tablestd{0.7} \\ 
\maclr{} & $39.2$\tablestd{0.7} & 	$17.3$\tablestd{0.5} & 	$50.9$\tablestd{1.0} & 	$38.6$\tablestd{0.6} & 	$19.7$\tablestd{0.1} & 	$56.4$\tablestd{0.8} \\
\maclrdescs{} & $38.5$\tablestd{0.4} & $17.0$\tablestd{0.5} & $49.1$\tablestd{0.8} & $36.3$\tablestd{0.4} & $18.3$\tablestd{0.7} & $52.4$\tablestd{0.6} \\
\groov{} & $17.5$\tablestd{0.6} & $4.2$\tablestd{0.2} & $9.4$\tablestd{0.4} & 	$11.4$\tablestd{0.8} & $4.3$\tablestd{0.4} & $	9.1$\tablestd{0.9} \\
\groovdescs{} & $1.1$\tablestd{0.2} & $0.4$\tablestd{0.2} & $1.3$\tablestd{0.3} & $4.0$\tablestd{0.3} & $0.9$\tablestd{0.1} & $1.9$\tablestd{0.4} \\
\lightxml{} & $12.5$\tablestd{1.9} & $6.3$\tablestd{0.9} & 	$19.5$\tablestd{2.9} & 	$7.5$\tablestd{0.7} &	$7.0$\tablestd{0.3} & 	$25.1$\tablestd{0.3} \\ \midrule
\rowcolor{gray!20} \multicolumn{7}{c}{\bf{5-shot}} \\ \midrule

\semsupxc{} & $\mathbf{63.3}$\tablestd{0.3} & $\mathbf{26.3}$\tablestd{0.3} & $\mathbf{67.6}$\tablestd{0.1} & $\mathbf{51.8}$\tablestd{0.1}  & $\mathbf{26.1}$\tablestd{0.2} & $\mathbf{70.0}$\tablestd{0.8}   \\ 
\maclr{} & $51.2$\tablestd{0.4} & $23.0$\tablestd{0.2}   & $63.8$\tablestd{0.7} & $42.4$\tablestd{0.3}    & $21.6$\tablestd{0.4}   & $61.4$\tablestd{0.1} \\
\maclrdescs{} & $52.5$\tablestd{0.4} & $22.9$\tablestd{0.5} & $62.1$\tablestd{0.4} & $39.3$\tablestd{0.5} & $19.9$\tablestd{0.3} & $57.4$\tablestd{0.6} \\
\groov{} & $43.1$\tablestd{1.0}   & $14.2$\tablestd{0.5} & $33.3$\tablestd{1.1} & $24.2$\tablestd{0.8}  & $9.7$\tablestd{0.6}  & $19.4$\tablestd{1.5} \\
\groovdescs{} & $6.0$\tablestd{0.6} & $1.3$\tablestd{0.4} & $3.5$\tablestd{0.6} & $17.6$\tablestd{0.5} & $3.5$\tablestd{0.2} & $9.5$\tablestd{0.3} \\
\lightxml{} & $52.7$\tablestd{0.1} & $23.7$\tablestd{0.0}   & $62.6$\tablestd{0.3} & $50.9$\tablestd{0.7} &	$24.7$\tablestd{0.4} &	$64.3$\tablestd{0.4} \\ \midrule

\rowcolor{gray!20} \multicolumn{7}{c}{\bf{10-shot}} \\ \midrule

\semsupxc{} & $\mathbf{68.5}$\tablestd{0.1} & $\mathbf{28.3}$\tablestd{0.5} & $\mathbf{71.9}$\tablestd{2.1} & $56.8$\tablestd{0.3} & $\mathbf{27.7}$\tablestd{0.4} & $\mathbf{69.8}$\tablestd{0.5} \\
\maclr{} & $56.1$\tablestd{0.1} & $25.4$\tablestd{0.0} & $69.4$\tablestd{0.1} & $43.9$\tablestd{0.2} & $22.3$\tablestd{0.0}   & $62.9$\tablestd{0.3} \\
\maclrdescs{} & $57.0$\tablestd{0.2} & $26.7$\tablestd{0.3} & $69.7$\tablestd{0.3} & $41.6$\tablestd{0.3} & $21.4$\tablestd{0.2} & $60.3$\tablestd{0.5} \\
\groov{} & $46.9$\tablestd{0.6} & $18.0$\tablestd{0.1}   & $43.2$\tablestd{0.2} & $29.6$\tablestd{0.4} & $12.8$\tablestd{0.1} & $29.3$\tablestd{0.4} \\
\groovdescs{} & $10.1$\tablestd{0.4} & $2.0$\tablestd{0.1} & $4.8$\tablestd{0.4} &  $21.4$\tablestd{0.5} & $4.5$\tablestd{0.5} & $13.4$\tablestd{0.4} \\
\lightxml{} & $61.6$\tablestd{0.6} & $27.1$\tablestd{0.3} & $71.0$\tablestd{0.4} & $\mathbf{57.7}$\tablestd{0.3} & $27.5$\tablestd{0.3} & $69.3$\tablestd{0.7} \\ \midrule

\rowcolor{gray!20} \multicolumn{7}{c}{\bf{20-shot}} \\ \midrule

\semsupxc{} & $\mathbf{72.6}$\tablestd{0.2} & $\mathbf{30.8}$\tablestd{0.1} & $\mathbf{78.2}$\tablestd{0.4} & $61.7$\tablestd{0.5} & $\mathbf{29.9}$\tablestd{0.2} & $\mathbf{73.9}$\tablestd{0.3} \\ 
\maclr{}  & $59.0$\tablestd{0.3} & $27.2$\tablestd{0.0} & $73.2$\tablestd{0.2} & $47.6$\tablestd{0.2} & $24.1$\tablestd{0.1} & $66.9$\tablestd{0.3} \\
\maclrdescs{} & $60.6$\tablestd{0.5} & $27.4$\tablestd{0.4} & $73.2$\tablestd{0.4} & $47.2$\tablestd{0.3} & $23.5$\tablestd{0.2} & $66.3$\tablestd{0.2} \\
\groov{}  & $53.3$\tablestd{1.3} & $21.5$\tablestd{0.8} & $52.7$\tablestd{2.2} & $35.7$\tablestd{0.2} & $15.6$\tablestd{0.2} & $39.5$\tablestd{0.3} \\
\groovdescs{} & $16.4$\tablestd{0.3} & $3.6$\tablestd{0.4} & $10.0$\tablestd{0.6} & $29.0$\tablestd{0.3} & $6.4$\tablestd{0.4} & $18.1$\tablestd{0.4} \\ 
\lightxml{} & $67.8$\tablestd{0.5} & $30.4$\tablestd{0.1} & $76.6$\tablestd{0.0} & $\mathbf{63.1}$\tablestd{0.2} & $29.8$\tablestd{0.3} & $73.6$\tablestd{0.3} \\ 

% \semsupxc{} & $63.3$\tablestd{0.3} & $26.3$\tablestd{0.3} & $67.6$\tablestd{0.1} &  &  &  \\ 
% \maclr{} & $49.3$\tablestd{2.2} & $23.0$\tablestd{0.2}   & $63.8$\tablestd{0.7} &  &  &  \\
% \groov{} & $43.1$\tablestd{1.0} & $14.2$\tablestd{0.5} & $33.3$\tablestd{1.1} &  &  &  \\
% \lightxml{} & $52.7$\tablestd{0.1} & $23.7$\tablestd{0.0}   & $62.6$\tablestd{0.3} &  &  &  \\

% \tfidf{} & $44.8$ & $21.1$ & $58.1$ & $48.5$ & $27.4$ & $73.1$  \\

% \midrule

% \rowcolor{gray!20} \multicolumn{7}{c}{\bf{1-shot}} \\ \midrule
% \maclr{} & $44.8$ & $21.1$ & $58.1$ & $48.5$ & $27.4$ & $73.1$  \\
% \midrule

% \rowcolor{gray!20} \multicolumn{7}{c}{\bf{5-shot}} \\ \midrule
% \groov{} & $44.8$ & $21.1$ & $58.1$ & $48.5$ & $27.4$ & $73.1$  \\
% \midrule

% \rowcolor{gray!20} \multicolumn{7}{c}{\bf{10-shot}} \\ \midrule
% Light XML & $44.8$ & $21.1$ & $58.1$ & $48.5$ & $27.4$ & $73.1$  \\
% \midrule

% \rowcolor{gray!20} \multicolumn{7}{c}{\bf{20-shot}} \\ \midrule
% \tfidf{} & $44.8$ & $21.1$ & $58.1$ & $48.5$ & $27.4$ & $73.1$  \\
% \midrule

\bottomrule
\end{tabular}
% Resize
% }
\caption{
Detailed table for few-shot results. \semsupxc{} outperforms all other baselines with significant margins for $k = 1, 5, \& 10$ shot settings. For 20-shot we perform almost at par with fully supervised method of \lightxml{}, which otherwise performs poorly for zero-shot and lower values of k in few shot setting. 
% \pranjal{Please fill and move it to one side and place the plots on the other side.}
        }
\label{table:few_shot}
\end{table*}

%% file: LaTeX/Appendix_files/004_computatioal_efficiency.tex
\section{Computational Efficiency}
\label{app:computational_efficiency}

Extreme Classification necessitates that the models scale well in terms of time and memory efficiency with labels at both train and test times. \semsupxc{} uses contrastive learning for efficiency at train time. During inference, \semsupxc{} predicts on top 1000 shortlists by \tfidf{}, thereby achieving sub-linear time. Further, contextualized tokens for label descriptions are computed only once and stored in memory-mapped files, thus decreasing computational time significantly. Overall, our computational complexity can be represented by $\mathcal{O}(T_{IE} * N + T_{OE} * |Y| + k * N * T_{lex})$, where $T_{IE}$, $T_{OE}$ represent the time taken by input encoder and output encoder respectively, $N$ is the total number of input documents, |Y| is the number of all labels, k indicates the shortlist size and $|T_{lex}|$ denotes the time in soft-lexical computation between contextualized tokens of documents and labels. In our experiments, $T_{IE} * N >> T_{OE} * |Y|$ and $T_{IE} \approx T_{lex} *k$. Thus effectively, computational complexity is approximately equal to $\mathcal{O(T_{IE} * N)}$, which is in comparison to other SOTA extreme classification methods. 

To ensure efficiency at inference time, similar to training, \semsupxc{} predicts on top of 1000 labels shortlisted by \tfidf{}. Table~\ref{table:computation-stats} shows that \semsupxc{}'s throughput is comparable to deep baselines (\maclr{} and \groov{}) while demonstrating much better performance.
While \zest{} is significantly faster, \semsupxc{}'s \precision{1} is $2\times$ higher.
While \semsupxc{}'s storage is higher, $17$ GB of space on modern-day hard drives is trivial, especially given that the dataset has over a million labels.
\semsupxc{}'s \relaxed{} module requires contextualized representations for every token in the description, which contributes to the majority of the storage.
This shows that \semsupxc{} provides the best throughput-performance trade-off while having practical storage requirements.

% \input{LaTeX/Tables/computation_table.tex}

%% file: LaTeX/Appendix_files/006_oracle_table.tex
% \begin{table}[h]
% \centering
% \caption{Performance Metrics}
% \begin{tabular}{lcccccc}
% \hline
% Method & \textbf{Eurlex P@1} & \textbf{Eurlex R@10} & \textbf{Amazon P@1} & \textbf{Amazon R@10} & \textbf{Wiki P@1} & \textbf{Wiki R@10} \\
% \hline
% \zest{} & $84.9$ & $60.2$ & $87.6$ & $54.2$ & $26.3$ & $17.2$ \\
% \maclr{} & $60.7$ & $55.2$ & $46.0$ & $46.9$ & $28.0$ & $32.7$ \\
% \groov{} & $84.1$ & $49.4$ & $87.4$ & $47.9$ & $31.4$ & $29.0$ \\
% \semsupxc{} & $87.0$ & $62.9$ & $88.6$ & $71.6$ & $33.7$ & $34.1$ \\
% $Oracle_{\text{Seen}}$ & $\textbf{97.2}$ & $60.0$ & $\textbf{95.0}$ & $54.9$ & $\textbf{66.1}$ & $43.2$ \\
% \hline
% \end{tabular}
% \end{table}

% Table 2: Extreme Zero-shot Learning (EZ-XMC) comparison of different unsupervised methods.
\begin{table*}[t]
% \ifx\subformat\iclrformat
\centering
\resizebox{\dimexpr\linewidth }{!}{%
\begin{tabular}{lcccccc}
\toprule
Method & \textbf{Eurlex P@1} & \textbf{Eurlex R@10} & \textbf{Amazon P@1} & \textbf{Amazon R@10} & \textbf{Wiki P@1} & \textbf{Wiki R@10} \\

\midrule
\zest{} & $84.9$ & $60.2$ & $87.6$ & $54.2$ & $26.3$ & $17.2$ \\
\maclr{} & $60.7$ & $55.2$ & $46.0$ & $46.9$ & $28.0$ & $32.7$ \\
\groov{} & $84.1$ & $49.4$ & $87.4$ & $47.9$ & $31.4$ & $29.0$ \\
\semsupxc{} & $87.0$ & $\textbf{62.9}$ & $88.6$ & $\textbf{71.6}$ & $33.7$ & $34.1$ \\
$Oracle_{\text{Seen}}$ & $\textbf{97.2}$ & $60.0$ & $\textbf{95.0}$ & $54.9$ & $\textbf{66.1}$ & $\textbf{43.2}$ \\
\hline
\end{tabular}}

\caption{
Comparison of $Oracle_{Seen}$ on \gzsxc{} split. $Oracle_{Seen}$ is able to get high scores, despite completely ignoring unseen labels. However \semsupxc{} is the only method that  beats $Oracle_{Seen}$ by significant margins on \eurlex{} and \amazon{} datasets.
}
\label{tab:oracle_table}
\end{table*}

%% file: LaTeX/Tables/007_d_unseen_table.tex
% \begin{table}[h]
% \centering
% \caption{Performance Metrics}
% \begin{tabular}{lcccccc}
% \hline
% Method & \textbf{Eurlex P@1} & \textbf{Eurlex R@10} & \textbf{Amazon P@1} & \textbf{Amazon R@10} & \textbf{Wiki P@1} & \textbf{Wiki R@10} \\
% \hline
% \zest{} & $84.9$ & $60.2$ & $87.6$ & $54.2$ & $26.3$ & $17.2$ \\
% \maclr{} & $60.7$ & $55.2$ & $46.0$ & $46.9$ & $28.0$ & $32.7$ \\
% \groov{} & $84.1$ & $49.4$ & $87.4$ & $47.9$ & $31.4$ & $29.0$ \\
% \semsupxc{} & $87.0$ & $62.9$ & $88.6$ & $71.6$ & $33.7$ & $34.1$ \\
% $Oracle_{\text{Seen}}$ & $\textbf{97.2}$ & $60.0$ & $\textbf{95.0}$ & $54.9$ & $\textbf{66.1}$ & $43.2$ \\
% \hline
% \end{tabular}
% \end{table}

% Table 2: Extreme Zero-shot Learning (EZ-XMC) comparison of different unsupervised methods.
\begin{table}[t]
% \ifx\subformat\iclrformat
\centering
\resizebox{\dimexpr\linewidth }{!}{%
\begin{tabular}{lcc}

% \begin{tabular}{lcccccc}
\toprule
Method & $P(D_{unseen}, Y)@1$ & $R(D_{unseen}, Y)@10$ \\

\midrule
\groov{} & $0.0$ & $0.8$ \\
\zest{} & $0.2$ & $8.8$ \\
\maclr{} & $16.4$ & $23.7$ \\
\semsupxc{} & $\textbf{31.2}$ & $\textbf{36.8}$ \\
\hline
\end{tabular}}

\caption{
Comparison of $P(D_{unseen}, Y)$ and $R(D_{unseen}, Y)$ on \gzsxc{} split on \eurlex{} dataset. \groov{} and \zest{} perform very poorly demonstrating they are deriving high scores on \gzsxc{} solely from seen labels. \maclr{} performs decently, while \semsupxc{} performs the best indicating the contribution of unseen labels in it's high scores in \gzsxc{} setting.
}
\label{tab:d_unseen_table}
\end{table}

%% file: LaTeX/Appendix_files/05_qualitative_analysis.tex
\section{Qualitative Analysis}
\label{app:qual_analysis}

Table~\ref{tab:app_ablation_qual} shows multiple qualitative examples for which our model outperforms the next baseline \maclr{}. The examples were chosen so as to increase diversity of input document's topic, number of correct predictions and relative improvement over baseline. All examples are on \amazon{} dataset.   

\begin{table*}[t]
\centering
\resizebox{\linewidth}{!}{%
\begin{tabular}{cll}
\toprule
\multirow{2}{*}{\bf{Input Document}} & \multicolumn{2}{c}{\bf{Top 5 Predictions}} \\
\cmidrule{2-3}
& \multicolumn{1}{c}{\textbf{\semsupxcbig{}}} &  \multicolumn{1}{c}{\textbf{\maclr{}}} \\ 
\midrule
\multirow{5}{0.7\linewidth}{Start-Up: A Technician's Guide. In addition to being an excellent stand-alone self-instructional guide, ISA recommends this book to prepare for the Start-Up Domain of CCST Level I, II, and III examinations.} & test preparation & vocational tests \\
& schools \& teaching & graduate preparation \\
& \textbf{new} & test prep \& study guides \\
& \textbf{used and rental textbooks} &  testing\\
& software & vocational  \\
\midrule
\multirow{5}{0.7\linewidth}{Homecoming (High Risk Books). When Katey Bruscke's bus arrives in her unnamed hometown, she finds the scenery blurred, "as if my hometown were itself surfacing from beneath a black ocean." At the conclusion of new novelist Gussoff's "day-in-the-life-of" first-person narrative, the reader feels equally blurred by the relentless \ldots}
& \textbf{literature \& fiction} & friendship \\
& thriller \& suspense & mothers \& children\\
& thrillers &  drugs \\
& genre fiction & coming of age\\
& general & braille \\

\midrule
\multirow{5}{0.7\linewidth}{Rolls RM65 MixMax 6x4 Mixer. The new RM65b HexMix is a single rack space unit featuring 6 channels of audio mixing, each with an XLR Microphone Input and 1/4\" unbalanced Line Input. A unique feature of the 1/4\" line inputs is they may be internally reconfigured to operate as Inserts for the Microphone Input. Each channel, in \ldots}
& \textbf{studio recording equipment} & powered mixers\\
& powered mixers & hand mixers \\
& home audio & mixers \& accessories \\
& \textbf{musical instruments} & mixers \\
& speaker parts \& components & mixer parts \\

\midrule
\multirow{5}{0.7\linewidth}{Political Business in East Asia (Politics in Asia). The book offers a valuable analysis of the ties between politics and business in various East Asian countries..Pacific Affairs, Fall 2003}
& \textbf{international \& world politics} & international \\
& \textbf{business \& investing} & relations \\
& \textbf{politics \& social sciences} & policy \& current events \\
& \textbf{asian} & practical politics \\
& \textbf{economics} & international law \\

\midrule
\multirow{5}{0.7\linewidth}{Chicago Latrobe 550 Series Cobalt Steel Jobber Length Drill Bit Set with Metal Case, Gold Oxide Finish, 135 Degree Split Point, Wire Size, 60-piece, \#60 - \#1. This Chicago-Latrobe 550 series jobber length drill bit set contains 60 cobalt steel drill bits, including one each of wire gauge sizes \#60 through \#1, with a gold oxide finish and a \ldots}
& \textbf{drill bits} & industrial drill bits \\
& \textbf{twist drill bits} & step drill bits \\
& \textbf{power \& hand tools} & long length drill bits \\
& \textbf{jobber drill bits} & reduced shank drill bits \\
& \textbf{power tool accessories} & installer drill bits \\

\midrule
\multirow{5}{0.7\linewidth}{Raggedy Ann and Johnny Gruelle: A Bibliography of Published Works. Patricia Hall has written and lectured extensively on Gruelle and his contributions to American culture. Her collection of Gruelle's books, dolls, correspondence, original artwork, business records and photographs is one of the most comprehensive in the world. Many \ldots}
& \textbf{reference} & art \\
& history \& criticism & bibliographies \& indexes \\
& humor \& entertainment & art \& photography \\
& \textbf{publishing \& books} & arts \\
& \textbf{research \& publishing guides} & children's literature \\

\midrule
\multirow{5}{0.7\linewidth}{Harmonic Analysis and Applications (Studies in Advanced Mathematics). The present book may definitely be useful for anyone looking for particular results, examples, applications, exercises, or for a book that provides the skeleton for a good course on harmonic analysis. R. Brger; Monatsheft fr Mathematik; 127.1999.3}
& statistics & pure mathematics \\
& \textbf{science \& math} & algebra \\
& professional science & applied \\
& \textbf{new} & algebra \& trigonometry \\
& \textbf{used \& rental textbooks} & calculus \\

\midrule
\multirow{5}{0.7\linewidth}{Soldier Spies: Israeli Military Intelligence. After its brilliant successes in the Six-Day War, the War of Attrition and the campaign against Black September, A'MAN, Israel's oldest intelligence agency, fell prey to institutional hubris. A'MAN's dangerous overconfidence only deepened, Katz here reveals, after spectacular coups such as \ldots}
& israel & intelligence \& espionage \\
& middle east & espionage \\
& international \& world politics & national \& international security \\
& politics \& social sciences & middle eastern \\
& \textbf{history} & arms control \\

\midrule
\multirow{5}{0.7\linewidth}{SF Signature White Chocolate Fondue, 4-Pound (Pack of 2). Smooth and creamy SF Signature White Chocolate Fondue provides unparalleled flavor in an incredibly easy-to-use product. This white chocolate fondue works better than other fountain chocolates due to its low viscosity and great taste.  Packaged in two pound \ldots}
& chocolate & baking cocoa \\
& breads \& bakery & chocolate truffles \\
& pantry staples & chocolate assortments \\
& canned \& jarred food & baking chocolates \\
& \textbf{kitchen \& dining} & chocolate \\

\midrule
\multirow{5}{0.7\linewidth}{Little Monsters: Monster Friends and Family (1000). Kind of a demented Monsters, Inc. meets Oliver Twist, the 1989 live-action film Little Monsters takes every child's nightmare of a monster under the bed and spins it into a dark tale of a secret underworld where children and adults turn into monsters and run wild without rules or any \ldots}
& \textbf{movies \& tv} & movies \\
& film & \textbf{movies \& tv} \\
& musical genres & \textbf{tv} \\
& rock & film \\
& \textbf{tv} & theater \\

\midrule
\multirow{5}{0.7\linewidth}{adidas Women's Ayuna Sandal,Newnavy/Wht/Altitude,5 M. adidas is a name that stands for excellence in all sectors of sport around the globe. The vision of company founder Adolf Dassler has become a reality, and his corporate philosophy has been the guiding principle for successor generations. The idea was as simple as it was \ldots}
& girls & \textbf{sport sandals} \\
& clothing & shoes \\
& \textbf{sandals} & athletic \\
& sneakers & mountaineering boots \\
& outdoor & \textbf{sandals} \\

\bottomrule

\end{tabular}
}
\caption{
\label{table:app_qualitative}
Examples of class predictions from \semsupxc{} (our model) compared to MACLR~\citep{Xiong2022ExtremeZL}. Bold represents correct predictions.
}

\label{tab:app_ablation_qual}
\end{table*}

%% file: LaTeX/Appendix_files/009_label_sources.tex
\subsection{Alternate Sources of Descriptions}
\label{app:desc_sources}

We further assess the impact of utilizing class descriptions from various sources. In Table~\ref{tab:desc_sources}, we compare the performance of descriptions obtained from 1.) \textbf{Language Model generated:} We generate descriptions using variant of GPT-3 with 6.7B parameters, 2.) \textbf{Knowledge Base}: We use definitions provided in WordNet as descriptions, and 3.) \semsupxc{}: Our proposed method.
Specifically, we sample 20 descriptions from GPT-3, using the prompt ``List 20 diverse descriptions for <class\_name> in the context of the <legal/e-commerce> domain''. We set the temperature at 0.7 to ensure diversity of scraped descriptions. We do not evaluate the method on \wikipedia{} due to the large cost associated with scraping descriptions for more than 1 million labels. 
For sourcing descriptions from WordNet, we find the closest synset to class description, and use the corresponding definitions of the synset.

Our method consistently outperforms the others, with a 2 \precision{1} point improvement on \eurlex{} and a 1 \precision{1} point improvement on \amazon{}. Furthermore, unlike LLM-based approaches, our method can efficiently scale to datasets containing millions of labels, such as \wikipedia{}. \semsupxc{} generates more diverse descriptions compared to those available in WordNet and is also applicable to classes containing proper nouns or non-dictionary words.

%% file: main.bbl
\begin{thebibliography}{53}
\providecommand{\natexlab}[1]{#1}
\providecommand{\url}[1]{\texttt{#1}}
\expandafter\ifx\csname urlstyle\endcsname\relax
  \providecommand{\doi}[1]{doi: #1}\else
  \providecommand{\doi}{doi: \begingroup \urlstyle{rm}\Url}\fi

\bibitem[Agrawal et~al.(2013)Agrawal, Gupta, Prabhu, and
  Varma]{Agrawal2013MultilabelLW}
Agrawal, R., Gupta, A., Prabhu, Y., and Varma, M.
\newblock Multi-label learning with millions of labels: recommending advertiser
  bid phrases for web pages.
\newblock \emph{Proceedings of the 22nd international conference on World Wide
  Web}, 2013.

\bibitem[Babbar \& Sch{\"o}lkopf(2017)Babbar and
  Sch{\"o}lkopf]{Babbar2017DiSMECDS}
Babbar, R. and Sch{\"o}lkopf, B.
\newblock Dismec: Distributed sparse machines for extreme multi-label
  classification.
\newblock \emph{Proceedings of the Tenth ACM International Conference on Web
  Search and Data Mining}, 2017.

\bibitem[Bai et~al.(2009)Bai, Weston, Grangier, Collobert, Sadamasa, Qi,
  Chapelle, and Weinberger]{bai2009supervised}
Bai, B., Weston, J., Grangier, D., Collobert, R., Sadamasa, K., Qi, Y.,
  Chapelle, O., and Weinberger, K.
\newblock Supervised semantic indexing.
\newblock In \emph{Proceedings of the 18th ACM conference on Information and
  knowledge management}, pp.\  187--196, 2009.

\bibitem[Bengio et~al.(2019)Bengio, Dembczynski, Joachims, Kloft, and
  Varma]{bengio2019extreme}
Bengio, S., Dembczynski, K., Joachims, T., Kloft, M., and Varma, M.
\newblock Extreme classification (dagstuhl seminar 18291).
\newblock In \emph{Dagstuhl Reports}, volume~8. Schloss
  Dagstuhl-Leibniz-Zentrum fuer Informatik, 2019.

\bibitem[Bhatia et~al.(2015)Bhatia, Jain, Kar, Varma, and
  Jain]{Bhatia2015SparseLE}
Bhatia, K., Jain, H., Kar, P., Varma, M., and Jain, P.
\newblock Sparse local embeddings for extreme multi-label classification.
\newblock In \emph{NIPS}, 2015.

\bibitem[Chalkidis et~al.(2019)Chalkidis, Fergadiotis, Malakasiotis, and
  Androutsopoulos]{Chalkidis2019LargeScaleMT}
Chalkidis, I., Fergadiotis, M., Malakasiotis, P., and Androutsopoulos, I.
\newblock Large-scale multi-label text classification on eu legislation.
\newblock In \emph{ACL}, 2019.

\bibitem[Chang et~al.(2019)Chang, Yu, Zhong, Yang, and Dhillon]{Chang2019AMD}
Chang, W.-C., Yu, H.-F., Zhong, K., Yang, Y., and Dhillon, I.~S.
\newblock A modular deep learning approach for extreme multi-label text
  classification.
\newblock \emph{ArXiv}, abs/1905.02331, 2019.

\bibitem[Chang et~al.(2020)Chang, Yu, Zhong, Yang, and Dhillon]{xtransformer}
Chang, W.-C., Yu, H.-F., Zhong, K., Yang, Y., and Dhillon, I.~S.
\newblock Taming pretrained transformers for extreme multi-label text
  classification.
\newblock In \emph{Proceedings of the 26th ACM SIGKDD international conference
  on knowledge discovery \& data mining}, pp.\  3163--3171, 2020.

\bibitem[Dahiya et~al.(2021{\natexlab{a}})Dahiya, Agarwal, Saini, Gururaj,
  Jiao, Singh, Agarwal, Kar, and Varma]{dahiya2021siamesexml}
Dahiya, K., Agarwal, A., Saini, D., Gururaj, K., Jiao, J., Singh, A., Agarwal,
  S., Kar, P., and Varma, M.
\newblock Siamesexml: Siamese networks meet extreme classifiers with 100m
  labels.
\newblock In \emph{International Conference on Machine Learning}, pp.\
  2330--2340. PMLR, 2021{\natexlab{a}}.

\bibitem[Dahiya et~al.(2021{\natexlab{b}})Dahiya, Saini, Mittal, Shaw, Dave,
  Soni, Jain, Agarwal, and Varma]{dahiya2021deepxml}
Dahiya, K., Saini, D., Mittal, A., Shaw, A., Dave, K., Soni, A., Jain, H.,
  Agarwal, S., and Varma, M.
\newblock Deepxml: A deep extreme multi-label learning framework applied to
  short text documents.
\newblock In \emph{Proceedings of the 14th ACM International Conference on Web
  Search and Data Mining}, pp.\  31--39, 2021{\natexlab{b}}.

\bibitem[Dauphin et~al.(2014)Dauphin, T{\"u}r, Hakkani-T{\"u}r, and
  Heck]{dauphin2014zero}
Dauphin, Y.~N., T{\"u}r, G., Hakkani-T{\"u}r, D., and Heck, L.~P.
\newblock Zero-shot learning and clustering for semantic utterance
  classification.
\newblock In \emph{ICLR (Poster)}, 2014.

\bibitem[Devlin et~al.(2019{\natexlab{a}})Devlin, Chang, Lee, and
  Toutanova]{devlinbert}
Devlin, J., Chang, M., Lee, K., and Toutanova, K.
\newblock {BERT:} pre-training of deep bidirectional transformers for language
  understanding.
\newblock In Burstein, J., Doran, C., and Solorio, T. (eds.), \emph{Proceedings
  of the 2019 Conference of the North American Chapter of the Association for
  Computational Linguistics: Human Language Technologies, {NAACL-HLT} 2019,
  Minneapolis, MN, USA, June 2-7, 2019, Volume 1 (Long and Short Papers)}, pp.\
   4171--4186. Association for Computational Linguistics, 2019{\natexlab{a}}.
\newblock \doi{10.18653/v1/n19-1423}.
\newblock URL \url{https://doi.org/10.18653/v1/n19-1423}.

\bibitem[Devlin et~al.(2019{\natexlab{b}})Devlin, Chang, Lee, and
  Toutanova]{Devlin2019BERTPO}
Devlin, J., Chang, M.-W., Lee, K., and Toutanova, K.
\newblock Bert: Pre-training of deep bidirectional transformers for language
  understanding.
\newblock In \emph{NAACL}, 2019{\natexlab{b}}.

\bibitem[Formal et~al.(2021)Formal, Piwowarski, and
  Clinchant]{Formal2021SPLADESL}
Formal, T., Piwowarski, B., and Clinchant, S.
\newblock Splade: Sparse lexical and expansion model for first stage ranking.
\newblock \emph{Proceedings of the 44th International ACM SIGIR Conference on
  Research and Development in Information Retrieval}, 2021.

\bibitem[Friedland(2013)]{profanity}
Friedland, B.
\newblock profanity: A python library to check for (and clean) profanity in
  strings, 2013.
\newblock URL \url{https://github.com/ben174/profanity}.

\bibitem[Gao et~al.(2021{\natexlab{a}})Gao, Dai, and Callan]{coil}
Gao, L., Dai, Z., and Callan, J.
\newblock {COIL:} revisit exact lexical match in information retrieval with
  contextualized inverted list.
\newblock In Toutanova, K., Rumshisky, A., Zettlemoyer, L.,
  Hakkani{-}T{\"{u}}r, D., Beltagy, I., Bethard, S., Cotterell, R.,
  Chakraborty, T., and Zhou, Y. (eds.), \emph{Proceedings of the 2021
  Conference of the North American Chapter of the Association for Computational
  Linguistics: Human Language Technologies, {NAACL-HLT} 2021, Online, June
  6-11, 2021}, pp.\  3030--3042. Association for Computational Linguistics,
  2021{\natexlab{a}}.
\newblock \doi{10.18653/v1/2021.naacl-main.241}.
\newblock URL \url{https://doi.org/10.18653/v1/2021.naacl-main.241}.

\bibitem[Gao et~al.(2021{\natexlab{b}})Gao, Yao, and Chen]{Gao2021SimCSESC}
Gao, T., Yao, X., and Chen, D.
\newblock Simcse: Simple contrastive learning of sentence embeddings.
\newblock \emph{ArXiv}, abs/2104.08821, 2021{\natexlab{b}}.

\bibitem[Grandury(2021)]{spamdet}
Grandury, M.
\newblock roberta-base-finetuned-sms-spam-detection, 2021.
\newblock URL
  \url{https://huggingface.co/mariagrandury/roberta-base-finetuned-sms-spam-detection}.

\bibitem[Gupta et~al.(2021)Gupta, Bohra, Prabhu, Purohit, and
  Varma]{Gupta2021GeneralizedZE}
Gupta, N., Bohra, S., Prabhu, Y., Purohit, S., and Varma, M.
\newblock Generalized zero-shot extreme multi-label learning.
\newblock \emph{Proceedings of the 27th ACM SIGKDD Conference on Knowledge
  Discovery \& Data Mining}, 2021.

\bibitem[Hadsell et~al.(2006)Hadsell, Chopra, and
  LeCun]{hadsell2006dimensionality}
Hadsell, R., Chopra, S., and LeCun, Y.
\newblock Dimensionality reduction by learning an invariant mapping.
\newblock In \emph{2006 IEEE Computer Society Conference on Computer Vision and
  Pattern Recognition (CVPR'06)}, volume~2, pp.\  1735--1742. IEEE, 2006.

\bibitem[Hanjie et~al.(2022)Hanjie, Deshpande, and
  Narasimhan]{Hanjie2022SemanticSE}
Hanjie, A.~W., Deshpande, A., and Narasimhan, K.
\newblock Semantic supervision: Enabling generalization over output spaces.
\newblock \emph{ArXiv}, abs/2202.13100, 2022.

\bibitem[Jain et~al.(2019)Jain, Balasubramanian, Chunduri, and
  Varma]{Jain2019SliceSL}
Jain, H., Balasubramanian, V., Chunduri, B., and Varma, M.
\newblock Slice: Scalable linear extreme classifiers trained on 100 million
  labels for related searches.
\newblock \emph{Proceedings of the Twelfth ACM International Conference on Web
  Search and Data Mining}, 2019.

\bibitem[Jiang et~al.(2021)Jiang, Wang, Sun, Yang, Zhao, and
  Zhuang]{Jiang2021LightXMLTW}
Jiang, T., Wang, D., Sun, L., Yang, H., Zhao, Z., and Zhuang, F.
\newblock Lightxml: Transformer with dynamic negative sampling for
  high-performance extreme multi-label text classification.
\newblock In \emph{AAAI}, 2021.

\bibitem[Khandagale et~al.(2020)Khandagale, Xiao, and
  Babbar]{Khandagale2020BonsaiDA}
Khandagale, S., Xiao, H., and Babbar, R.
\newblock Bonsai: diverse and shallow trees for extreme multi-label
  classification.
\newblock \emph{Machine Learning}, pp.\  1 -- 21, 2020.

\bibitem[Larochelle et~al.(2008)Larochelle, Erhan, and
  Bengio]{larochelle2008zero}
Larochelle, H., Erhan, D., and Bengio, Y.
\newblock Zero-data learning of new tasks.
\newblock In \emph{AAAI}, volume~1, pp.\ ~3, 2008.

\bibitem[Lee et~al.(2019)Lee, Chang, and Toutanova]{InverseCloze}
Lee, K., Chang, M.-W., and Toutanova, K.
\newblock Latent retrieval for weakly supervised open domain question
  answering.
\newblock \emph{ArXiv}, abs/1906.00300, 2019.

\bibitem[Lee et~al.(2022)Lee, Ippolito, Nystrom, Zhang, Eck, Callison-Burch,
  and Carlini]{Lee2022DeduplicatingTD}
Lee, K., Ippolito, D., Nystrom, A., Zhang, C., Eck, D., Callison-Burch, C., and
  Carlini, N.
\newblock Deduplicating training data makes language models better.
\newblock In \emph{ACL}, 2022.

\bibitem[Lin et~al.(2014)Lin, Ding, Hu, and Wang]{Lin2014MultilabelCV}
Lin, Z., Ding, G., Hu, M., and Wang, J.
\newblock Multi-label classification via feature-aware implicit label space
  encoding.
\newblock In \emph{ICML}, 2014.

\bibitem[Liu et~al.(2017)Liu, Chang, Wu, and Yang]{Liu2017DeepLF}
Liu, J., Chang, W.-C., Wu, Y., and Yang, Y.
\newblock Deep learning for extreme multi-label text classification.
\newblock \emph{Proceedings of the 40th International ACM SIGIR Conference on
  Research and Development in Information Retrieval}, 2017.

\bibitem[Loshchilov \& Hutter(2019)Loshchilov and
  Hutter]{Loshchilov2019DecoupledWD}
Loshchilov, I. and Hutter, F.
\newblock Decoupled weight decay regularization.
\newblock In \emph{ICLR}, 2019.

\bibitem[McAuley \& Leskovec(2013)McAuley and Leskovec]{McAuley2013HiddenFA}
McAuley, J. and Leskovec, J.
\newblock Hidden factors and hidden topics: understanding rating dimensions
  with review text.
\newblock \emph{Proceedings of the 7th ACM conference on Recommender systems},
  2013.

\bibitem[Medini et~al.(2019)Medini, Huang, Wang, Mohan, and
  Shrivastava]{medini2019extreme}
Medini, T. K.~R., Huang, Q., Wang, Y., Mohan, V., and Shrivastava, A.
\newblock Extreme classification in log memory using count-min sketch: A case
  study of amazon search with 50m products.
\newblock \emph{Advances in Neural Information Processing Systems}, 32, 2019.

\bibitem[Mittal et~al.(2021)Mittal, Dahiya, Agrawal, Saini, Agarwal, Kar, and
  Varma]{mittal2021decaf}
Mittal, A., Dahiya, K., Agrawal, S., Saini, D., Agarwal, S., Kar, P., and
  Varma, M.
\newblock Decaf: Deep extreme classification with label features.
\newblock In \emph{Proceedings of the 14th ACM International Conference on Web
  Search and Data Mining}, pp.\  49--57, 2021.

\bibitem[Nam et~al.(2016)Nam, Menc{\'\i}a, and F{\"u}rnkranz]{nam2016all}
Nam, J., Menc{\'\i}a, E.~L., and F{\"u}rnkranz, J.
\newblock All-in text: Learning document, label, and word representations
  jointly.
\newblock In \emph{Thirtieth AAAI Conference on Artificial Intelligence}, 2016.

\bibitem[Pappas \& Henderson(2019)Pappas and Henderson]{pappas2019gile}
Pappas, N. and Henderson, J.
\newblock Gile: A generalized input-label embedding for text classification.
\newblock \emph{Transactions of the Association for Computational Linguistics},
  7:\penalty0 139--155, 2019.

\bibitem[Prabhu et~al.(2018)Prabhu, Kag, Harsola, Agrawal, and Varma]{Parabel}
Prabhu, Y., Kag, A., Harsola, S., Agrawal, R., and Varma, M.
\newblock Parabel: Partitioned label trees for extreme classification with
  application to dynamic search advertising.
\newblock pp.\  993--1002, 04 2018.
\newblock ISBN 978-1-4503-5639-8.
\newblock \doi{10.1145/3178876.3185998}.

\bibitem[Raffel et~al.(2019)Raffel, Shazeer, Roberts, Lee, Narang, Matena,
  Zhou, Li, and Liu]{t5_google}
Raffel, C., Shazeer, N., Roberts, A., Lee, K., Narang, S., Matena, M., Zhou,
  Y., Li, W., and Liu, P.~J.
\newblock Exploring the limits of transfer learning with a unified text-to-text
  transformer, 2019.
\newblock URL \url{https://arxiv.org/abs/1910.10683}.

\bibitem[Rajaee \& Pilehvar(2021)Rajaee and Pilehvar]{Rajaee2021ACAisotropic}
Rajaee, S. and Pilehvar, M.~T.
\newblock A cluster-based approach for improving isotropy in contextual
  embedding space.
\newblock In \emph{ACL}, 2021.

\bibitem[Reimers et~al.(2019)Reimers, Gurevych, and
  .]{Reimers2019SentenceBERTSE}
Reimers, N., Gurevych, I., and .
\newblock Sentence-bert: Sentence embeddings using siamese bert-networks.
\newblock \emph{ArXiv}, abs/1908.10084, 2019.

\bibitem[Rios \& Kavuluru(2018)Rios and Kavuluru]{rios2018few}
Rios, A. and Kavuluru, R.
\newblock Few-shot and zero-shot multi-label learning for structured label
  spaces.
\newblock In \emph{Proceedings of the Conference on Empirical Methods in
  Natural Language Processing. Conference on Empirical Methods in Natural
  Language Processing}, volume 2018, pp.\  3132. NIH Public Access, 2018.

\bibitem[Simig et~al.(2022)Simig, Petroni, Yanki, Popat, Du, Riedel, and
  Yazdani]{Simig2022OpenVE}
Simig, D., Petroni, F., Yanki, P., Popat, K., Du, C., Riedel, S., and Yazdani,
  M.
\newblock Open vocabulary extreme classification using generative models.
\newblock \emph{ArXiv}, abs/2205.05812, 2022.

\bibitem[Thakur et~al.(2021)Thakur, Reimers, Ruckl'e, Srivastava, and
  Gurevych]{Thakur2021BEIRAH}
Thakur, N., Reimers, N., Ruckl'e, A., Srivastava, A., and Gurevych, I.
\newblock Beir: A heterogenous benchmark for zero-shot evaluation of
  information retrieval models.
\newblock \emph{ArXiv}, abs/2104.08663, 2021.

\bibitem[Turc et~al.(2019)Turc, Chang, Lee, and Toutanova]{Turc2019WellReadSL}
Turc, I., Chang, M.-W., Lee, K., and Toutanova, K.
\newblock Well-read students learn better: On the importance of pre-training
  compact models.
\newblock \emph{arXiv: Computation and Language}, 2019.

\bibitem[Wang et~al.(2018)Wang, Li, Wang, Zhang, Shen, Zhang, Henao, and
  Carin]{wang2018joint}
Wang, G., Li, C., Wang, W., Zhang, Y., Shen, D., Zhang, X., Henao, R., and
  Carin, L.
\newblock Joint embedding of words and labels for text classification.
\newblock In \emph{Proceedings of the 56th Annual Meeting of the Association
  for Computational Linguistics (Volume 1: Long Papers)}, pp.\  2321--2331,
  2018.

\bibitem[Wei \& Zou(2019)Wei and Zou]{eda}
Wei, J. and Zou, K.
\newblock Eda: Easy data augmentation techniques for boosting performance on
  text classification tasks, 2019.
\newblock URL \url{https://arxiv.org/abs/1901.11196}.

\bibitem[Williams et~al.(2018)Williams, Nangia, and Bowman]{MNLI}
Williams, A., Nangia, N., and Bowman, S.
\newblock A broad-coverage challenge corpus for sentence understanding through
  inference.
\newblock In \emph{Proceedings of the 2018 Conference of the North American
  Chapter of the Association for Computational Linguistics: Human Language
  Technologies, Volume 1 (Long Papers)}, pp.\  1112--1122. Association for
  Computational Linguistics, 2018.
\newblock URL \url{http://aclweb.org/anthology/N18-1101}.

\bibitem[Wydmuch et~al.(2018)Wydmuch, Jasinska, Kuznetsov, Busa-Fekete, and
  Dembczynski]{Wydmuch2018ANG}
Wydmuch, M., Jasinska, K., Kuznetsov, M., Busa-Fekete, R., and Dembczynski, K.
\newblock A no-regret generalization of hierarchical softmax to extreme
  multi-label classification.
\newblock In \emph{NeurIPS}, 2018.

\bibitem[Xiong et~al.(2022)Xiong, Chang, Hsieh, Yu, and
  Dhillon]{Xiong2022ExtremeZL}
Xiong, Y., Chang, W.-C., Hsieh, C.-J., Yu, H.-F., and Dhillon, I.~S.
\newblock Extreme zero-shot learning for extreme text classification.
\newblock \emph{ArXiv}, abs/2112.08652, 2022.

\bibitem[Xue et~al.(2021)Xue, Constant, Roberts, Kale, Al-Rfou, Siddhant,
  Barua, and Raffel]{Xue2021mT5AM}
Xue, L., Constant, N., Roberts, A., Kale, M., Al-Rfou, R., Siddhant, A., Barua,
  A., and Raffel, C.
\newblock mt5: A massively multilingual pre-trained text-to-text transformer.
\newblock In \emph{NAACL}, 2021.

\bibitem[Yen et~al.(2017)Yen, Huang, Dai, Ravikumar, Dhillon, and
  Xing]{Yen2017PPDsparseAP}
Yen, I. E.-H., Huang, X., Dai, W., Ravikumar, P., Dhillon, I.~S., and Xing,
  E.~P.
\newblock Ppdsparse: A parallel primal-dual sparse method for extreme
  classification.
\newblock \emph{Proceedings of the 23rd ACM SIGKDD International Conference on
  Knowledge Discovery and Data Mining}, 2017.

\bibitem[You et~al.(2019)You, Zhang, Wang, Dai, Mamitsuka, and
  Zhu]{You2019AttentionXMLLT}
You, R., Zhang, Z., Wang, Z., Dai, S., Mamitsuka, H., and Zhu, S.
\newblock Attentionxml: Label tree-based attention-aware deep model for
  high-performance extreme multi-label text classification.
\newblock In \emph{NeurIPS}, 2019.

\bibitem[Zhang et~al.(2021)Zhang, Chang, Yu, and Dhillon]{zhang2021fast}
Zhang, J., Chang, W.-C., Yu, H.-F., and Dhillon, I.
\newblock Fast multi-resolution transformer fine-tuning for extreme multi-label
  text classification.
\newblock \emph{Advances in Neural Information Processing Systems},
  34:\penalty0 7267--7280, 2021.

\bibitem[Zhang et~al.(2022)Zhang, Shen, Wu, Xie, Hao, Wang, Wang, and
  Han]{zhang2022metadata}
Zhang, Y., Shen, Z., Wu, C.-H., Xie, B., Hao, J., Wang, Y.-Y., Wang, K., and
  Han, J.
\newblock Metadata-induced contrastive learning for zero-shot multi-label text
  classification.
\newblock In \emph{Proceedings of the ACM Web Conference 2022}, pp.\
  3162--3173, 2022.

\end{thebibliography}
